\documentclass[lettersize,journal]{IEEEtran}
\usepackage{amsmath,amsfonts}
\usepackage{algorithmic}
\usepackage{algorithm}
\usepackage{array}
\usepackage[caption=false,font=normalsize,labelfont=sf,textfont=sf]{subfig}
\usepackage{textcomp}
\usepackage{stfloats}
\usepackage{url}
\usepackage{verbatim}
\usepackage{graphicx}
\usepackage{cite}
\usepackage{color,soul}
\makeatletter 
\newcount\SOUL@minus
\makeatother  
\usepackage{booktabs}
\usepackage[hidelinks]{hyperref}
\usepackage{epstopdf}
\usepackage{amsmath,amsthm} 
\usepackage{amsfonts} 
\usepackage{graphicx} 
\begin{document}

\title{A Survey of Imitation Learning: Algorithms, Recent Developments, and Challenges}

\author{Maryam~Zare$^*$,~Parham~M.~Kebria,~\IEEEmembership{Member,~IEEE,}
	Abbas~Khosravi,~\IEEEmembership{Senior Member,~IEEE,}\\
	and~Saeid~Nahavandi,~\IEEEmembership{Fellow,~IEEE}
\thanks{Maryam Zare, Parham M. Kebria, and Abbas Khosravi are with the Institute for Intelligent Systems Research and Innovation (IISRI), Deakin University, Waurn Ponds, 3216 VIC, Australia ($^*$Corresponding author: {\tt\footnotesize{mzare@deakin.edu.au}}).}
\thanks{Saeid Nahavandi is with the Institute for Intelligent Systems Research and Innovation (IISRI), Deakin University, Waurn Ponds, 3216 VIC, Australia, and also with the Harvard Paulson School of Engineering and Applied Sciences, Harvard University, Allston, MA 02134 USA.\newline This work has been submitted to the IEEE for possible publication. Copyright may be transferred without notice, after which this version may no longer be accessible.}
}

\markboth{}%
{Shell \MakeLowercase{\textit{et al.}}: A Sample Article Using IEEEtran.cls for IEEE Journals}


\maketitle

\begin{abstract}

In recent years, the development of robotics and artificial intelligence (AI) systems has been nothing short of remarkable. As these systems continue to evolve, they are being utilized in increasingly complex and unstructured environments, such as autonomous driving, aerial robotics, and natural language processing. As a consequence, programming their behaviors manually or defining their behavior through reward functions (as done in reinforcement learning (RL)) has become exceedingly difficult. This is because such environments require a high degree of flexibility and adaptability, making it challenging to specify an optimal set of rules or reward signals that can account for all possible situations. In such environments, learning from an expert's behavior through imitation is often more appealing. This is where imitation learning (IL) comes into play - a process where desired behavior is learned by imitating an expert's behavior, which is provided through demonstrations.

This paper aims to provide an introduction to IL and an overview of its underlying assumptions and approaches. It also offers a detailed description of recent advances and emerging areas of research in the field. Additionally, the paper discusses how researchers have addressed common challenges associated with IL and provides potential directions for future research. Overall, the goal of the paper is to provide a comprehensive guide to the growing field of IL in robotics and AI.
\end{abstract}


\section{Introduction}
Traditionally, machines and robots have been manually programmed to learn autonomous behavior \cite{osa2018algorithmic}. Traditional methods require experts to provide specific, hard-coded rules regarding the actions that a machine must perform, as well as the characteristics of the environment in which the machine operates. However, developing such rules requires considerable time and coding expertise \cite{ravichandar2020recent}. In order to automate the tedious manual hard-coding of every behavior, a learning approach is required \cite{schaal1999imitation}. Imitation learning provides an avenue for teaching the desired behavior by demonstrating it. IL techniques have the potential to reduce the problem of teaching a task to that of providing demonstrations, thus eliminating the need for explicit programming or the development of task-specific reward functions \cite{schaal1999imitation}. The concept of IL is based on the premise that human experts are capable of demonstrating the desired behavior even when they are unable to program it into a machine or robot. As such, IL can be leveraged in any system that requires autonomous behavior similar to that of a human expert \cite{osa2018algorithmic}.

The main purpose of IL is to enable agents to learn to perform a specific task or behavior by imitating an expert through the provision of demonstrations \cite{deng2018learning}. Demonstrations are used to train learning agents to perform a task by learning a mapping between observations and actions. By utilizing IL, agents are able to transition from repeating simple predetermined behaviors in constrained environments to taking optimal autonomous actions in unstructured environments, without imposing too much burden on the expert \cite{ravichandar2020recent}. As a result, IL approaches have the potential to offer significant benefits to a wide range of industries, including manufacturing \cite{zhu2018robot}, health care \cite{van2010superhuman}, autonomous vehicles \cite{le2022survey,jalali2019optimal}, and the gaming industry \cite{aytar2018playing}. In these applications, IL allows subject-matter experts, who may not possess coding skills or knowledge of the system, to program autonomous behavior in machines or robots efficiently. Although the idea of learning by imitation has been around for some time, recent achievements in computing and sensing, along with a growing demand for artificial intelligence applications, have increased the significance of IL \cite{ogenyi2019physical, sun2019survey}. Consequently, the number of publications in the field has increased significantly in recent years.

\begin{figure*}[!t]
\centering
\includegraphics[width=7in]{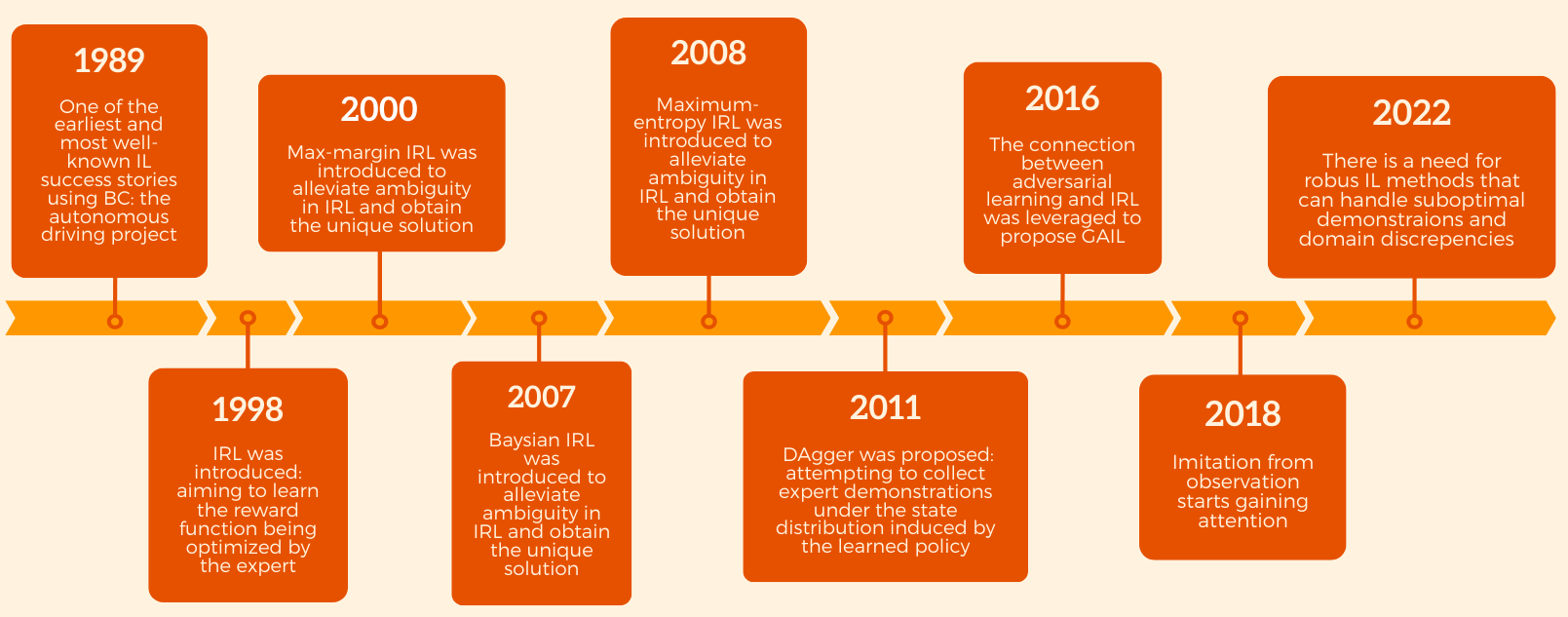}
\caption{A historical timeline of IL research illustrating key achievements in the field.}
\hfil
\label{timeline}
\end{figure*}

Multiple surveys of IL have been published over the past two decades, each focusing on different aspects of the field's development (Fig. \ref{timeline}). Schaal \cite{schaal1999imitation} presented the first survey of IL, focusing on IL as a route to create humanoid robots. More recently, Osa et al. \cite{osa2018algorithmic} provided an algorithmic perspective on IL, while Hussein et al. \cite{hussein2017imitation} provided a comprehensive review of the design options for each stage of the IL process. Most recently, Le Mero et al. \cite{le2022survey} provided a comprehensive overview of IL-based techniques for end-to-end autonomous driving systems. 

Despite the existence of a large number of surveys on IL, a new survey is necessary to capture the latest advances in this rapidly evolving field and provide an up-to-date overview of the state of the art. With the field gaining increasing interest and having diverse applications, a comprehensive survey could serve as an essential reference for newcomers, as well as provide an overview of different use cases. We acknowledge that IL is a constantly evolving field, with new algorithms, techniques, and applications being developed. Therefore, our survey aims to consolidate the vast amount of research on IL, making it easier for researchers and practitioners to navigate. Moreover, we aim to identify gaps and challenges in the current research, providing a clear direction for future work. Lastly, we aim to make the concepts and techniques of IL more accessible to a wider audience, including researchers from related fields, to enhance the understanding of this area. Overall, we strongly believe that our survey will make significant contributions to advancing the field of IL and guide future research in this exciting area.

The objective of this survey paper is to present a comprehensive overview of the field of IL. To achieve this, we will organize our discussion of IL approaches based on historical and logical reasons. Initially, we will introduce the two broad categories of approaches to IL: behavioral cloning (BC) and inverse reinforcement learning (IRL). We will discuss their formulations, developments, strengths, and limitations. Additionally, we will explore how adversarial imitation learning (AIL) extends IRL by introducing an adversarial context to the learning process. We will underscore the benefits of integrating adversarial training into IL and assess the current progress in the AIL field. Furthermore, we will introduce imitation from observation (IfO) as a novel technique that aims to learn from state-only (no actions) demonstrations. We will discuss the significance of IfO and how it incorporates and extends the previous categories of BC, IRL, and AIL in different methods to tackle the challenge of learning from state-only observations. Finally, we will discuss the challenges that IL techniques encounter in real-world scenarios, such as sub-optimal demonstrations and domain discrepancies between the expert and the learner. We will conclude with a discussion of the different approaches to IL, their limitations, and future research directions that can be taken to address them.

\section{Behavioral Cloning}

BC is an IL technique that treats the problem of learning a behavior as a supervised learning task \mbox{\cite{pomerleau1991efficient, ross2011reduction}}. BC involves training a model to mimic an expert's behavior by learning to map the state of the environment to the corresponding expert action. The expert's behavior is recorded as a set of state-action pairs, also known as demonstrations. During the training process, the model is provided with these demonstrations as inputs and is trained to learn a function that maps the current state to the corresponding expert action. Once the model is trained, it can use the learned function to generate actions for new states that it has not encountered before.

One advantage of BC is that it requires no knowledge of the underlying dynamics of the environment \mbox{\cite{pomerleau1991efficient}}. Instead, it relies solely on the provided demonstrations to learn the behavior. Additionally, BC is computationally efficient since it involves training a supervised learning model, which is a well-studied problem in machine learning.

\begin{figure}[!t]
\centering
\includegraphics[width=0.5\textwidth]{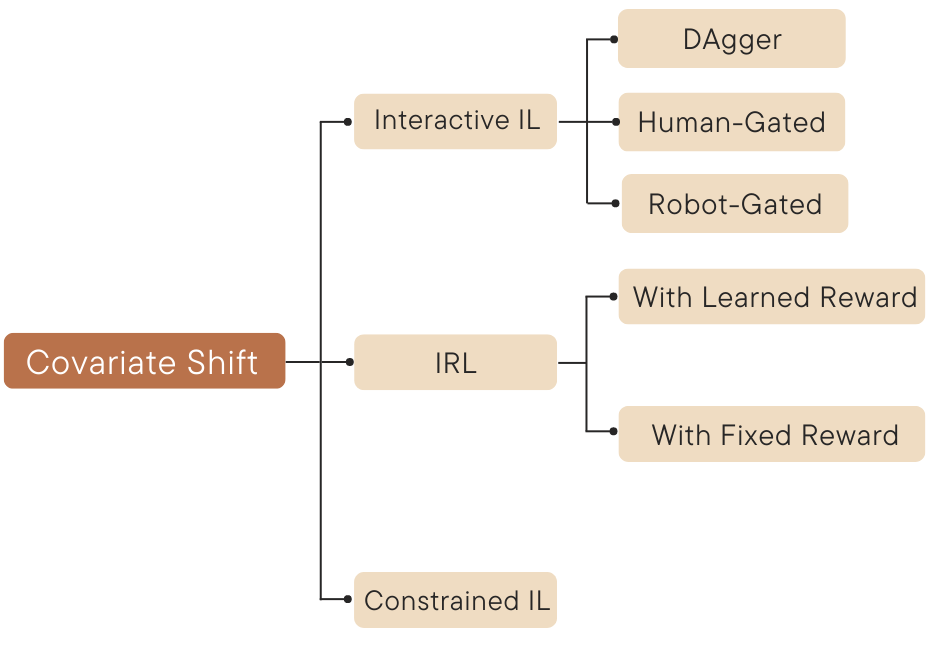}
\caption{A categorization of methods addressing the covariate shift problem. Interactive IL assumes access to an online expert. DAgger like algorithms require the expert to provide corrective labels for each action taken by the agent. On the other hand, human-gated and robot-gated methods provide corrective labels only when they are requested by the expert or agent, respectively. Unlike interactive IL, IRL methods do not require access to an online expert. These methods require an underlying RL algoirthm to optimize a reward function (either learned from demonstrations or fixed). Lastly, by constraining the agent to known regions of the space covered by demonstrations, constrained IL attempts to address covariate shift problems that cannot be expressed or solved using the other two categories.}
\hfil
\label{covshift}
\end{figure}

Despite its simplicity, the BC approach has a significant drawback - the covariate shift problem \cite{pomerleau1988alvinn}. This problem arises because during training, the learner is trained on states generated by the expert policy, but during testing, the learner is tested on states induced by its action \cite{zhou2022domain}. As a result, the state distribution observed during testing can differ from that observed during training. The problem with BC supervised approach is that the agent does not know how to return to the demonstrated states when it drifts and encounters out-of-distribution states \cite{reddy2020sqil}. Covariate shift is particularly dangerous in safety-critical situations such as driving \cite{roche2021multimodal}, as the agent may encounter novel situations that it has not seen during training, and its ability to recover from mistakes can be critical to avoid accidents. To address the covariate shift problem and improve the robustness of the BC approach, three broad research areas have been identified (Fig. \ref{covshift}).

The first and most popular area is interactive IL. Algorithms of this type are based on the assumption that the agent has access to an online expert who can be consulted during training. Dataset aggregation (DAgger) \cite{ross2011reduction} is the earliest interactive IL method and proposes to train the agent on its own state distribution to resolve the train and test time mismatch problem. DAgger queries the expert to relabel the data collected by the agent with the appropriate action that should have been taken. However, due to frequent queries, the human expert is subjected to a significant cognitive burden, resulting in inaccurate or delayed feedback that adversely affects the training process \cite{li2022efficient}. Consequently, determining when and how to engage human subjects is one of the key challenges of interactive IL algorithms \cite{hoque2021thriftydagger}. 

Rather than providing continuous feedback, “human-gated” interactive IL algorithms \cite{kelly2019hg,mandlekar2020human} extend DAgger to allow the expert to decide when to provide the corrective interventions. For example, human-gated DAgger (HG-DAgger) \cite{kelly2019hg} rolls out the agent trajectory until the expert determines that the agent has reached an unsafe region of the state space. In this case, the human expert intervenes by taking control of the system and guiding the agent back to a safe state. Using this method, no constraints limit the amount of human intervention. Li et al. \cite{li2022efficient} propose a method that learns to minimize human intervention and adaptively maximize automation during training. To accomplish this, when the human expert issues intervention, it incurs a cost to the agent, which the agent learns to minimize during its training process. 

However, the use of these algorithms depends on human experts constantly monitoring the agent to decide when to intervene, which imposes a significant burden on them.  To tackle this challenge, there has been an increasing interest in ``robot-gated" algorithms \cite{Zhang2017query,hoque2021lazydagger,menda2019ensembledagger,hoque2021thriftydagger} that allow robots to actively ask humans for intervention. For example, SafeDAgger \cite{Zhang2017query} uses an auxiliary safety policy, which determines the likelihood of the agent deviating from the expert's trajectory, as a signal for the agent to transfer control over to the expert. LazyDAgger \cite{hoque2021lazydagger} extends SafeDAgger to reduce the number of context switches between the expert and autonomous control. A recent robot-gated approach called ThriftyDAgger \cite{hoque2021thriftydagger} aims to intervene only when states are sufficiently novel (out-of-distribution) or risky (prone to result in task failure). In addition, the intervention burden is reduced by limiting the total number of interventions to a budget specified by the user.

The second area of research addressing the covariate shift problem consists of algorithms that estimate the support of the expert occupancy measure and then specify a reward that encourages the agent to remain on the support of the expert \cite{wang2019random,reddy2020sqil,brantley2019disagreement}. The reward function is then optimized using RL. Unlike interactive IL, these algorithms do not assume access to an online expert and only rely on demonstrations and further interactions with the environment. The most popular algorithms are based on IRL, which addresses the covariate shift by training an RL agent to consistently match the demonstrations over time. A detailed discussion of these methods will be provided in section III and IV. However, these methods often use complex and unstable approximation techniques involving adversarial training to learn a reward function from demonstrations \cite{reddy2020sqil,arjovsky2017towards}. In this section, a recent alternative line of research is reviewed that also employs RL, but instead of learning a reward function, it uses a simple fixed reward function. The key idea is to incentivize the agent to consistently remain on the support of the expert policy over time by encouraging it to return to demonstrated states when facing new states outside the expert’s support \cite{dadashi2020primal}.

Wang et al. \cite{wang2019random} estimate the support of the expert policy using a kernelized version of principal component analysis. The support estimation process produces a score that increases as a state-action pair moves closer to the support of the expert policy. The score is then used to construct an intrinsic reward function. 

Reddy et al. \cite{reddy2020sqil} propose soft Q IL (SQIL). SQIL encourages the agent to imitate the expert in demonstrated states by using an extremely sparse reward function – assigns a constant reward of +1 to transitions inside expert demonstrations and a constant reward of 0 to all other transitions. The reward function encourages the agent to return to demonstrated states after encountering out-of-distribution states. The proposed model outperforms simple BC and shows good performance even with a limited number of demonstrations.

Brantley et al. \cite{brantley2019disagreement} use expert demonstrations to train an ensemble of policies, with the variance of their predictions used as the cost. Inherently, the variance (cost) outside of expert support would be higher since ensemble policies will be more likely to disagree on states not seen in the demonstrations. An RL algorithm minimizes this cost in combination with a supervised BC cost. As a result, the RL cost assists the agent in returning to the expert distribution, whereas the supervised cost ensures that the agent mimics the expert within the expert's distribution.

Lastly, the third area of algorithms aims to constrain the agent to known regions of the space covered by the demonstrator without relying on an interactive expert or leveraging RL. These methods are particularly beneficial and practical for real-world applications where safety constraints must be met, such as in healthcare, autonomous driving, and industrial processes \cite{brantley2021expert}. In \cite{Bansal2O19ChauffeurNet}, the authors attempt to overcome the covariate shift problem in autonomous driving by augmenting the imitation loss with additional losses to discourage bad driving. Moreover, additional data is provided to the agent in the form of synthetic perturbations to the expert's trajectory. The perturbations expose the model to non-expert behavior such as collisions and provide an important signal for the added losses to avoid these behaviors. 

Wong et al. \cite{wong2022error} propose a learned error detection system to indicate when an agent is in a potential failure state. In this way, the error detector can constrain the policy to execute only on states seen previously in the demonstrations and prevent potentially unstable behavior by resetting the agent to a well-known configuration or terminating the execution. \cite{chang2021mitigating} considers an offline IL setting where the agent is provided with two datasets: one small dataset of expert policy state-action pairs and one large dataset of state-action-next state transitions from a potentially suboptimal behavior policy. Their solution to the covariate shift problem consists of training a dynamics model on the sub-optimal demonstrations and applying a high penalty in regions of the state-action space that are not well covered by the data.

\begin{figure}[!t]
\centering
\includegraphics[width=3.4in]{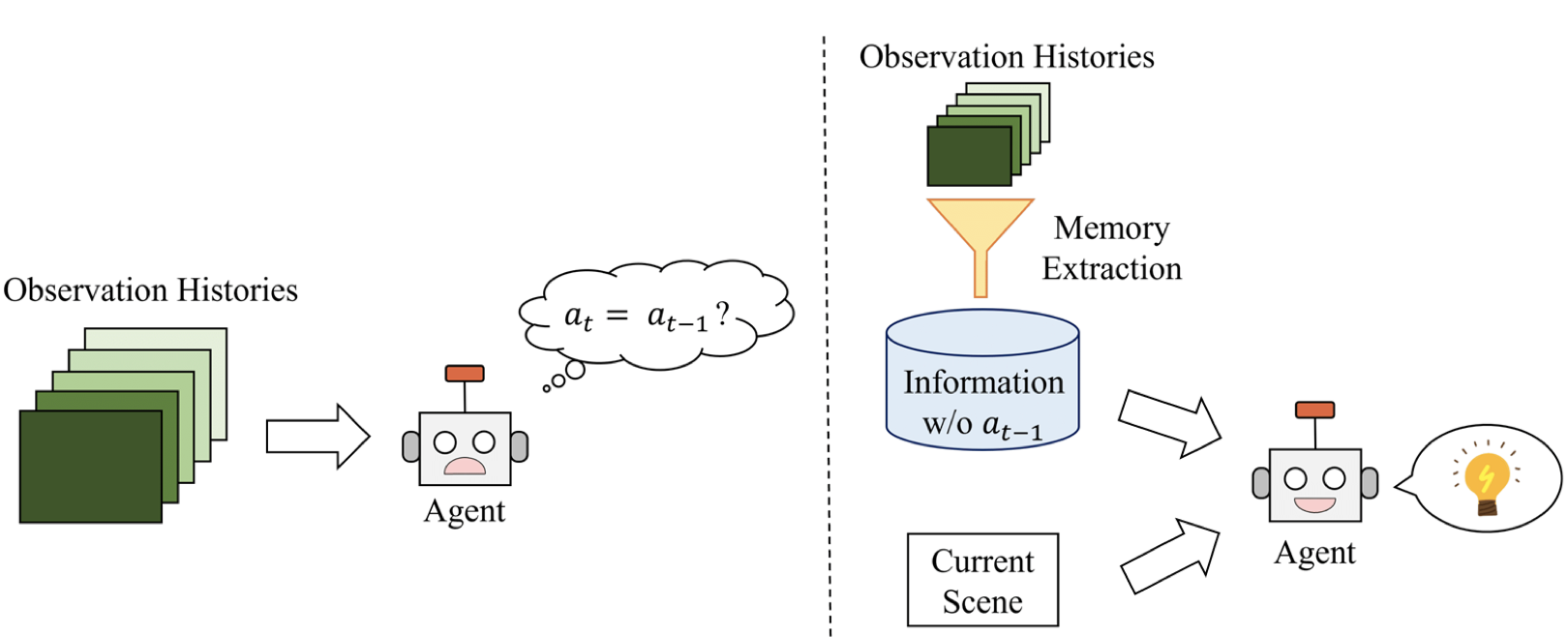}
\caption{Left: BC might learn a shortcut from prior observations that outputs the previous action as the current action. Right: a copycat-free memory extraction module. The shortcut is no longer available using historical information \cite{chuang2022resolving}.}
\label{copycat}
\end{figure}

It is difficult for the supervised learning approach used in behavioral cloned policies to identify the underlying causes of expert actions, leading to a phenomenon known as ``causal misidentification" \cite{de2019causal}. In the training procedure, the causal structure of the interaction between the expert and the environment is not taken into account. Therefore, cloned policies might fail to distinguish nuisance correlates from the true causes of expert actions \cite{de2019causal}. When training and testing distributions are the same, ignoring nuisance correlates might not pose a problem as they continue to hold in the test dataset. In BC, however, ignoring causality is particularly problematic because of the distributional shift \cite{de2019causal}. The causal structure of the underlying sequential decision process in IL further exacerbates this problem. This is because the causal structure–past actions influence future observations–often creates additional complex nuisance correlates. To address this problem, \cite{de2019causal} learns a mapping from causal graphs to policies and then uses targeted interventions–either expert queries or environment interactions–to find the optimal policy. Subsequent work by Wen et al. \cite{wen2020fighting} explores a prominent class of causal confusion problems known as “the copycat problem.” This problem happens when the expert actions are highly correlated over time. In this scenario, the agent learns to cheat by copying the expert’s previous actions. To address the copycat problem, \cite{wen2020fighting} proposes an adversarial approach for learning a feature representation that ignores the information about the known nuisance correlates–the previous actions–while retaining the necessary information to predict the next action. Chuang et al. \cite{chuang2022resolving} extend \cite{wen2020fighting} to high-dimensional image observations. Using a memory extraction module, they attempt to extract historical features from observation histories while removing as much information as possible regarding previous actions (Fig. \ref{copycat}).

Traditionally, BC involves training an explicit neural network \cite{bworld}. Unfortunately, conventional explicit models struggle to model discontinuities, making policies incapable of switching decisively between different behaviors. This issue arises because explicit models are unable to represent discontinuities with neural networks built with continuous activation functions, which is almost always the case with neural networks \cite{bworld}. In contrast, implicit models are capable of representing sharp discontinuities, despite only having continuous layers in the network \cite{bworld}.

The implicit BC model presented in \cite{florence2022implicit} turns BC into an energy-based modeling problem \cite{song2021train} by training a neural network that takes both observations and actions as input and outputs a single value that is low for expert actions and high for non-expert actions \cite{bworld}. A trained implicit BC policy selects the action input with the lowest score for a given observation. This method requires more computation than explicit BC models, both during training and during inference. However, the results demonstrate that it can often outperform traditional explicit baselines in robotic manipulation tasks, both in the real world and in simulation.

\section{Inverse Reinforcement
Learning}

In addition to behavioral cloning, another key approach to imitation learning is IRL \mbox{\cite{russell1998learning}}. IRL involves an apprentice agent that aims to infer the reward function underlying the observed demonstrations, which are assumed to come from an expert who acts optimally \mbox{\cite{piot2016bridging}}. Once the reward function is inferred, it is optimized to train an apprentice policy through RL \mbox{\cite{lian2021robust}}.

RL agents, unlike the agents in BC, learn by continually interacting with their environment, observing the consequences of their actions, and altering their behavior to maximize long-term cumulative reward \mbox{\cite{sutton2018reinforcement, nguyen2020deep}}. This process involves using reinforcement signals to learn the long-term consequences of each action, allowing the agent to recover from mistakes \mbox{\cite{brantley2019disagreement}}. Because of this capability, IRL is less sensitive to covariate shift compared to BC \mbox{\cite{ross2011reduction}}.

IRL has been widely used in a variety of applications, such as robotics manipulation, autonomous navigation, game playing, and natural language processing \mbox{\cite{abbeel2004apprenticeship, ho2016generative, ziebart2008maximum}}. Nonetheless, devising an effective IRL algorithm for learning from demonstrations is a challenging task, mainly due to two major reasons.

Firstly, IRL can be computationally expensive and resource-intensive. In part, this is due to the fact that the agent must interact repeatedly with its environment to accurately estimate the reward function \mbox{\cite{li2022efficient, torabi2019recent}}. Additionally, the nature of this process can be inherently unsafe, particularly when dealing with high-risk applications such as autonomous driving or aircraft control \mbox{\cite{kuderer2015learning}}.

Furthermore, a typical IRL approach follows an iterative process that involves alternating between reward estimation and policy training, which results in poor sample efficiency \mbox{\cite{ho2016generative, ng2000algorithms, ziebart2008maximum}}. Consequently, there has been significant research aimed at addressing these issues to enhance the sample efficiency of IRL algorithms while maintaining the safety and accuracy of the learned policy. Some of these approaches include methods that utilize human guidance to reduce the number of interactions required to estimate the reward function accurately \mbox{\cite{hadfield2016cooperative}}.

The second major challenge of IRL arises due to the inherent ambiguity in the relationship between the policy and the reward function. Specifically, a policy can be optimal with respect to an infinite number of reward functions \mbox{\cite{ng2000algorithms, kim2021reward}}. To address this challenge, researchers have proposed various methods to introduce additional structure into the reward function. There are roughly three categories of IRL methods that aim to address this ambiguity \mbox{\cite{jarrett2021inverse}}.

The first category is maximum-margin methods. The key idea in maximum-margin methods is to infer a reward function that explains the optimal policy more thoroughly than all other policies by a margin. These methods address the discussed ambiguity problem by converging on a solution that maximizes some margin. A foundational method in this category is the work of Ng et al. \cite{ng2000algorithms}. They estimate the reward function for which the given policy is optimal using a linear program while maximizing a margin. Another major work is maximum margin planning (MMP) \cite{ratliff2006maximum} which seeks to find a weighted linear mapping of features to rewards so that the estimated policy is “close” to the demonstrated behaviors. \cite{bagnell2006boosting,ratliff2009learning} build on and extend MMP to nonlinear hypothesis spaces by utilizing a family of functional gradient techniques. 

The adoption of feature-based reward functions gave rise to a variety of approaches making use of feature expectations for margin optimization. Abbeel and Ng \cite{abbeel2004apprenticeship} propose two foundational methods (max-margin and projection) for maximizing the feature expectation loss margin without assuming access to the expert’s policy. As with many other IRL methods, these methods have the drawback of limiting the agent's performance to the quality of the expert. To address this limitation, Syed and Schapire \cite{syed2007game} propose a game-theoretic approach capable of training a policy with superior performance to an expert.

The second category of IRL algorithms aims to solve the ambiguity problem by maximizing the entropy of the resulting policy. MaxEntIRL \cite{ziebart2008maximum} was the first IRL method to utilize maximum entropy. Ziebart's work \cite{ziebart2008maximum} demonstrates that the maximum entropy paradigm is capable of handling expert suboptimality and stochasticity by using the distribution over possible trajectories. Subsequent works \cite{aghasadeghi2011maximum, kalakrishnan2013learning} extend the MaxEntIRL algorithm  to continuous state-action spaces using path integral methods. Optimizing a full forward Markov Decision Process (MDP) iteratively becomes intractable in high-dimensional continuous state-action spaces. To overcome this complexity, these studies exploit the local optimality of demonstrated trajectories. 

In many prior methods, detailed features are manually extracted using domain knowledge, which can be linearly combined into a reward, such as the distance between the ball and the cup for a robotic ball-in-cup game \cite{boularias2011relative}. While linear representations are sufficient in many domains, they may be overly simplistic for complex real-world tasks, particularly when reward values are derived from raw sensory data. Wulfmeier et al. \cite{wulfmeier2015maximum} propose maximum entropy deep IRL, a generalization of MaxEntIRL that utilizes neural networks to model complex, nonlinear reward functions. In addition, instead of using pre-extracted features, the deep architecture is further extended to learn features through Convolution layers. This is an important step towards automating the learning process \cite{hussein2017imitation}. A further study by Fin et al. \cite{finn2016guided} proposes guided cost learning (GCL), which improves the sampling efficiency of \cite{wulfmeier2015maximum} since \cite{wulfmeier2015maximum} relies on a large number of expert transitions to estimate the reward function. GCL learns a nonlinear reward function in the inner loop of a policy optimization (in contrast to early IRL methods). This allows it to scale effectively to complex control problems. A further advantage of this method is that it utilizes the raw state of the system instead of predefined features to construct the reward function, which reduces the engineering burden. 

Bayesian algorithms constitute the third category of IRL algorithms. Methods under this category use the expert's actions as the evidence for updating the reward function estimate. A posterior distribution over candidate reward functions is derived from a prior distribution over the rewards and a likelihood of the reward hypothesis. Various models of likelihood have been proposed over the years. BIRL \cite{ramachandran2007bayesian} is the earliest Bayesian IRL technique, based on a Boltzmann distribution for modeling the likelihood. A variety of distributions can be used as the prior over the reward functions \cite{ramachandran2007bayesian}. For example, a Beta distribution is appropriate for planning problems with large rewards dichotomy. Analytically obtaining the posterior in the  continuous space of reward functions is extremely difficult. To address this issue, \cite{ramachandran2007bayesian} uses Markov chain Monte Carlo (MCMC) to derive a sample-based estimate of the posterior mean. Instead of computing the posterior mean, \cite{choi2011map} computes the maximum a posteriori (MAP) reward function. \cite{choi2011map} argues that the posterior mean is not the most suitable approach for reward inference as it integrates over the entire reward space, even those not consistent with the observed behavior, in the loss function. Levine et al. \cite{levine2011nonlinear} propose a Bayesian IRL algorithm that uses a nonlinear function of features to represent the reward. They use a Gaussian process prior on reward values and a kernel function to determine the structure of the reward. The kernel's hyperparameters are learned through the Bayesian GP framework, leading to the learning of the reward structure. 

\begin{figure*}[!t]
\centering
\includegraphics[width=7in]{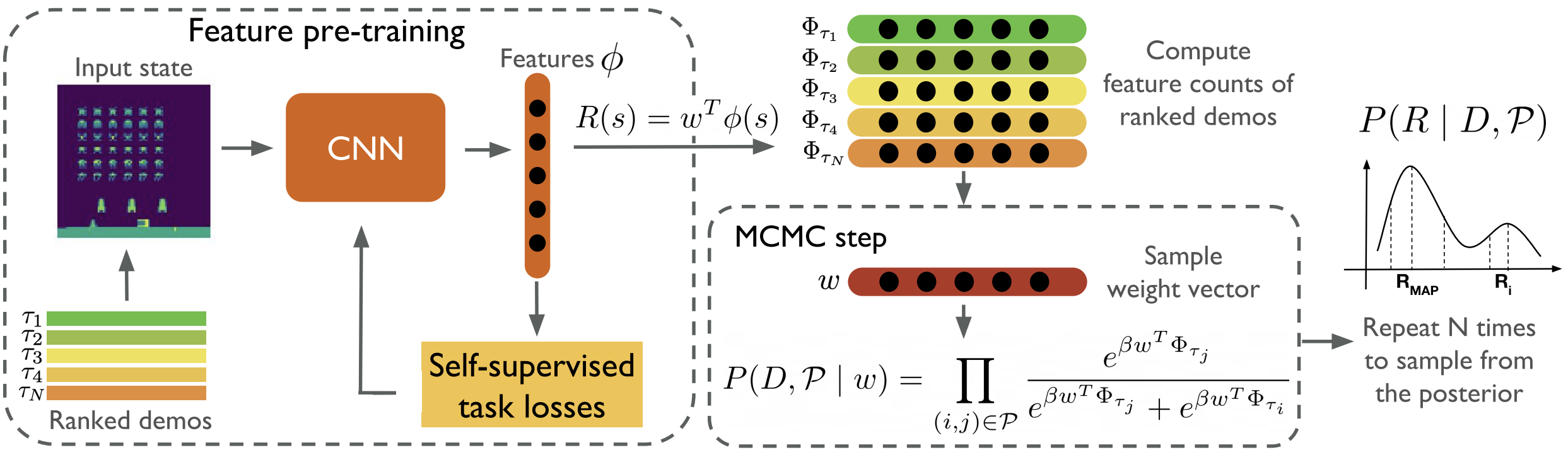}
\caption{A low-dimensional state feature embedding is pre-trained using ranked demonstrations \cite{brown2020safe}. A linear combination of learned features is used to derive the reward function. A pairwise ranking likelihood is used by MCMC proposal evaluations to estimate the likelihood of preferences over demonstrations given a proposal (\emph{w}). Utilizing the pre-computed embeddings of the ranked demonstrations makes MCMC sampling highly efficient; There is no need for data collection during inference or an MDP solver.}

\label{ranked}
\end{figure*}

Classical BIRL algorithms are computationally intractable in complex high-dimensional environments since generating each posterior sample requires solving an entire MDP. This limitation has prevented these methods from scaling beyond small tabular settings. To overcome this limitation, \cite{brown2020safe} generates samples from the posterior distribution without using an MDP solver by proposing an alternative likelihood formulation that leverages preference labels over demonstrations (Fig. \ref{ranked}). Another approach to address the scalability issue is approximate variational reward IL (AVRIL) \cite{chan2021scalable}. This approach simultaneously learns an imitating policy and an approximate posterior distribution over the reward in an offline setting. Unlike traditional sampling or MAP-based techniques, AVRIL relies on variational inference to estimate the posterior distribution accurately.

Most existing IRL algorithms rely on the unrealistic assumption that the transition model, and sometimes the expert's policy, are known beforehand \cite{ziebart2008maximum,ratliff2006maximum,levine2011nonlinear,ng2000algorithms}. However, in real-world scenarios, the agent often has to estimate the expert policy and transition dynamics from samples, leading to errors in the recovered reward function \cite{lindner2022active, metelli2021provably}. In their analysis, \cite{metelli2021provably} breaks down this error into components from estimating the expert policy and transition model. Based on their analysis (in a tabular setting), they propose an efficient sampling strategy focused on transferring the learned reward function to a fully known target environment. It is assumed, however, that the agent can query the transition dynamics for arbitrary states and actions. To remove this assumption, active exploration for IRL (AceIRL) \cite{lindner2022active} focuses on developing an efficient exploration strategy. This strategy aims to explore both the environment dynamics and the expert policy such that an arbitrary IRL algorithm can infer the reward function as effectively as possible. By utilizing previous observations, AceIRL builds confidence intervals that capture feasible reward functions and find exploration policies that prioritize the most relevant regions of the environment. 

\section{Adversarial Imitation Learning}

Scaling IRL algorithms to larger environments has been a major challenge despite their success in generating policies that replicate expert behavior \mbox{\cite{finn2016guided, ho2016model, levine2012continuous}}. This challenge arises due to the computational complexity of many IRL algorithms, which often require RL to be executed in an inner loop \mbox{\cite{ho2016generative}}. AIL offers a promising solution to the computational challenges of IRL by searching for the optimal policy without fully solving an RL sub-problem at each iteration \mbox{\cite{ho2016generative}}. AIL involves a two-player game between an agent and an adversary (discriminator) where the adversary attempts to distinguish agent trajectories from expert trajectories \mbox{\cite{deka2023arc}}. The agent, on the other hand, endeavors to deceive the adversary by generating trajectories that closely resemble expert trajectories. Through this adversarial process, the agent gradually improves its imitation of the expert's behavior until it converges to a policy that closely resembles the expert's policy. AIL has demonstrated statistically significant improvements over existing methods in multiple benchmark environments, including robotics, autonomous driving, and game playing \mbox{\cite{ho2016generative, 9990591, song2018multi}}.

The effectiveness of AIL in addressing the limitations of IRL has spurred continued research in this area. The first AIL method that gained prominence is known as generative AIL (GAIL) \mbox{\cite{ho2016generative}}. In GAIL, the reward function measures the ability of the agent to imitate the expert's behavior. To do this, GAIL utilizes a discriminator network trained to distinguish between the expert's behavior and the agent's generated trajectories. The reward signal is then derived from the confusion of the discriminator, reflecting how difficult it is to tell whether a given trajectory is generated by the agent or the expert. By maximizing this reward signal, the agent is incentivized to generate trajectories that closely resemble the expert's behavior. Over the years, numerous improvements have been proposed to the original algorithm to improve its sample efficiency, scalability, and robustness \mbox{\cite{orsini2021matters}}, including changes to the discriminator’s loss function \mbox{\cite{fu2017learning}} and switching from on-policy to off-policy agents \mbox{\cite{kostrikov2018discriminator}}.

In AIL, the objective is to enable the agent to generate trajectories that are similar to those of the expert. This involves the use of distance measures to quantify the similarity between the two. Different AIL methods employ various similarity measures to match the distribution over states and actions encountered by the agent with that of the expert \mbox{\cite{dadashi2020primal}}. For example, GAIL makes use of the Shannon-Jensen divergence, while some methods, such as AIRL \mbox{\cite{fu2017learning}}, use the Kullback-Leibler divergence. However, recent research by Arjovsky et al. \mbox{\cite{arjovsky2017wasserstein}} has shown that replacing f-divergences with the Wasserstein distance through its dual formulation can result in improved training stability, a technique that several AIL methods have implemented \mbox{\cite{li2017infogail,kostrikov2018discriminator}}. Given these developments, exploring new similarity measures holds the potential to discover novel AIL methods.

Most AIL methods, just like GANs (generative adversarial networks) \cite{goodfellow2014generative}, use a min-max optimization approach to minimize the distance between the state-action distributions of the expert and agent, while maximizing a reward signal derived from the discriminator's confusion. However, this approach can be challenging to train due to issues such as vanishing gradients and convergence failure \mbox{\cite{arjovsky2017towards}}. To overcome these challenges, methods such as primal wasserstein IL (PWIL) \mbox{\cite{dadashi2020primal}} have been developed, which approximates Wasserstein distances through a primal-dual approach.

\section{Imitation from Observation}

The prevailing paradigm in IL assumes that the learner has access to both states and actions demonstrated by an expert \cite{edwards2019imitating}. However, this often necessitates collecting data explicitly for IL purposes \cite{edwards2019imitating}. In robotics, for instance, the expert must teleoperate the robot or move its joints manually (kinesthetic learning) \cite{hu2022robot}, and in gaming, the expert may require a special software stack. In both cases, considerable operator expertise is required, and useful demonstrations are limited to those recorded under artificial conditions. These limiting factors have motivated recent efforts in IfO \cite{liu2018imitation}, where the expert's actions are unknown. In contrast to previous methods, imitation from observation is a more natural way to learn from experts and is more in tune with how humans and animals approach imitation in general. It is common for humans to learn new behaviors by observing other humans without being aware of their low-level actions (e.g., muscle commands). Humans learn a wide range of tasks, from weaving to swimming to playing games, by watching videos online. While there may be huge gaps in body shapes, sensing modalities, and timing, they show an incredible ability to apply the knowledge gained from online demonstrations \cite{aytar2018playing}. 

Enabling agents to learn from demonstrations without the action information makes a large number of previously inapplicable resources, such as videos on the Internet, available for learning \cite{ICML19a-torabi}. Additionally, it opens up the possibility of learning from agents with different embodiments whose actions are unknown or cannot be matched. The use of state-only demonstrations for IL is not new \cite{ijspeert2002movement}. However, recent deep learning and visual recognition developments \cite{choe2021indoor} have equipped researchers with more powerful tools to approach the problem, particularly when dealing with raw visual observations \cite{torabi2019recent}. 

\begin{figure}[!t]
\centering
\includegraphics[width=3in]{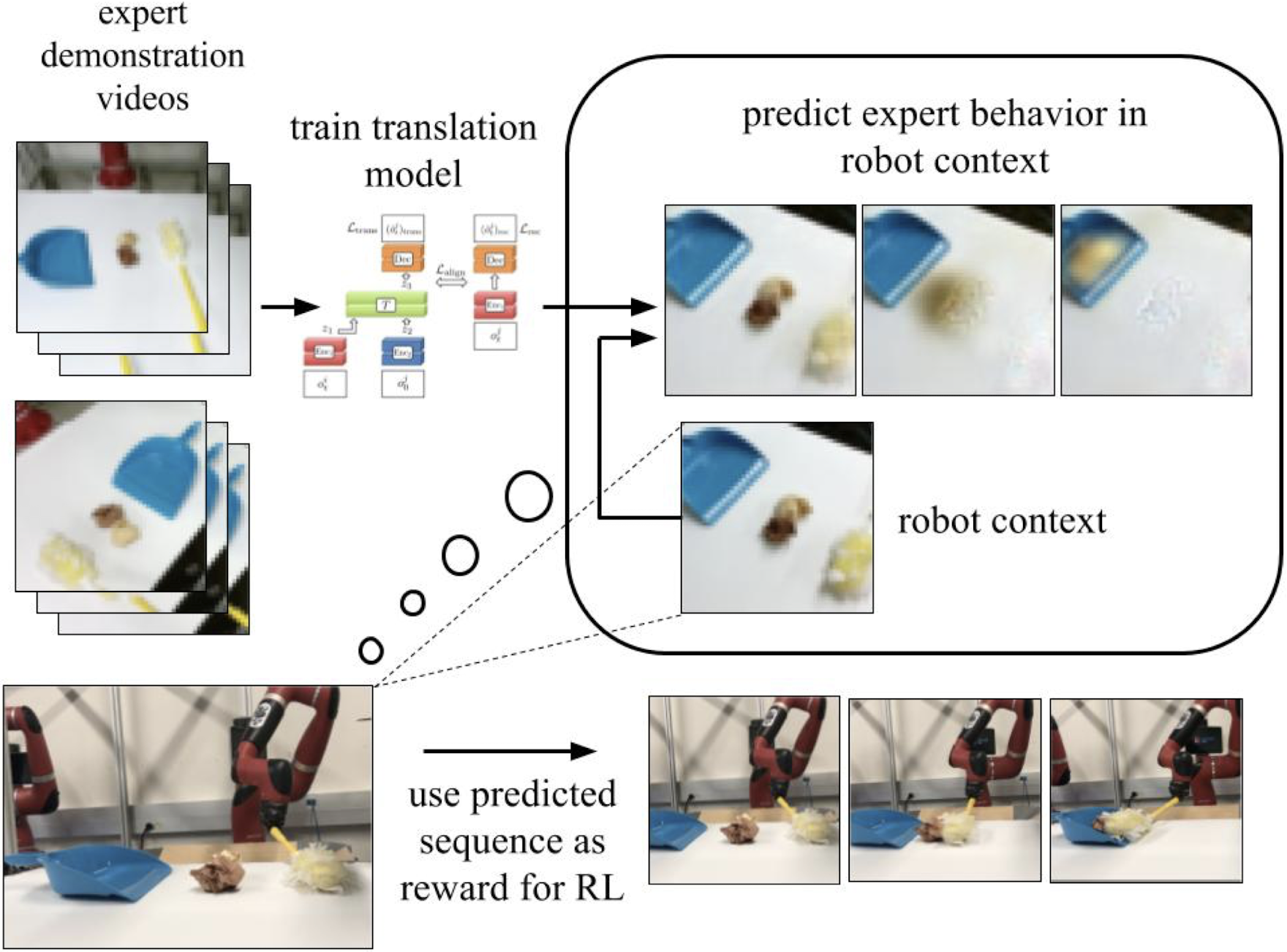}
\caption{A context translation model is trained on several videos of expert demonstrations \cite{liu2018imitation}. The robot observes the context of the task it must perform during the learning process. The model then determines what an expert would do in the context of the robot.}
\label{context}
\end{figure}

Liu et al. \cite{liu2018imitation} propose an imitation from observation method that learns an imitator policy from raw videos using context-aware translation. Their algorithm utilizes a context translation model that converts demonstrations from the expert’s context (e.g., a third-person viewpoint) to the agent’s context (e.g., a first-person viewpoint). The model is then used to predict the expert behavior in the context of the robot (Fig. \ref{context}). Using the predicted observations, a reward function is defined that is made up of a penalty for deviating from the expert’s translated features – encoded from input observations – and a penalty for encountering observations that are different from the translated observations. RL is then used to optimize the derived reward function. There are two drawbacks that limit the applicability of this method. First, it is assumed that demonstrations from different contexts are aligned in time which is rarely the case in the real world \cite{raychaudhuri2021cross}. Second, learning the translation model requires a large number of demonstrations \cite{liu2018imitation}. A further limitation is that it cannot address systematic domain shifts, such as differences in embodiment \cite{liu2018imitation}.

\begin{figure}[!t]
\centering
\includegraphics[width=3.4in]{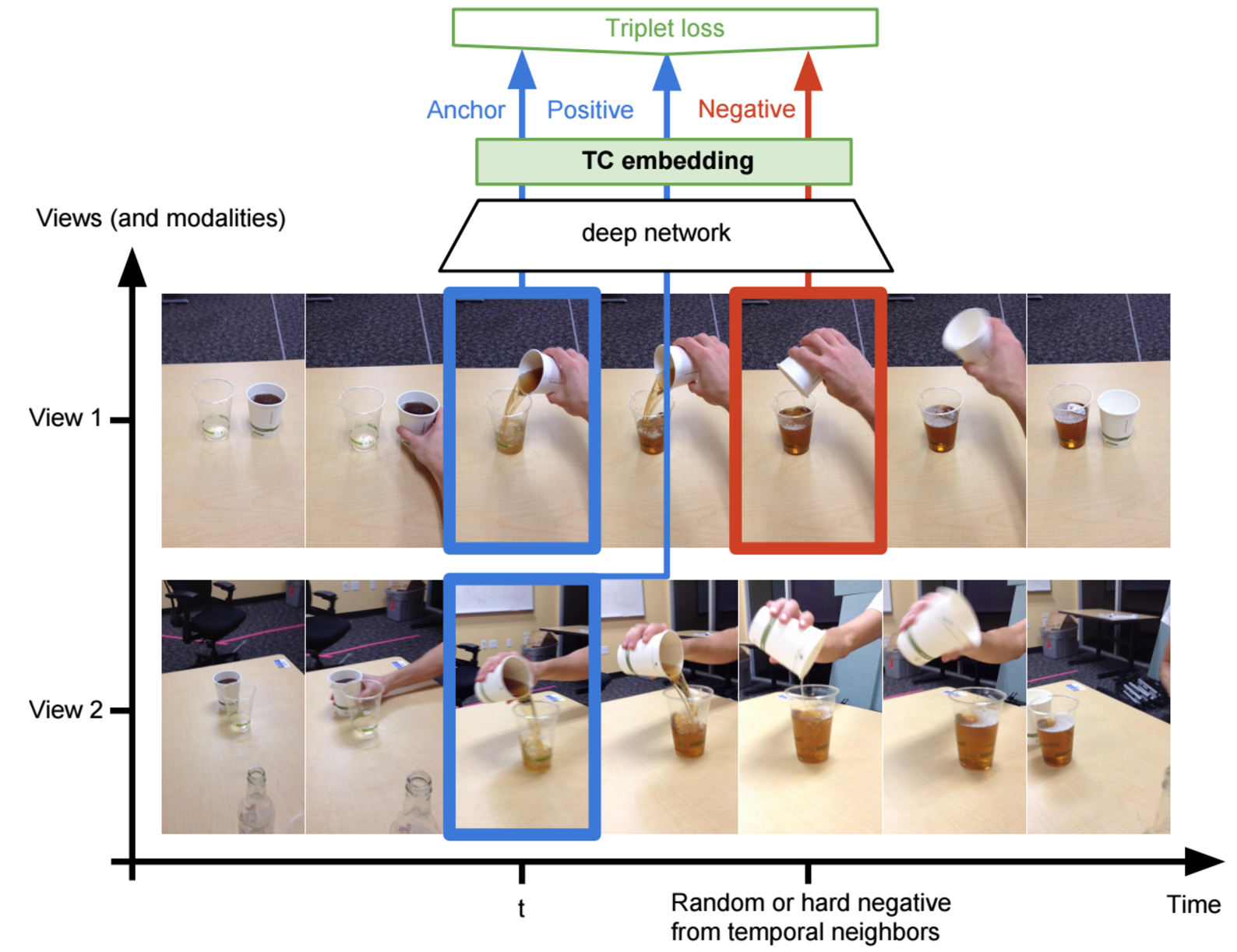}
\caption{The embedding space encourages co-occurring frames from different viewpoints to be in close proximity to each other, while images captured from the same viewpoint but at different times should be far apart \cite{sermanet2018time}.}

\label{embedding}
\end{figure}

Sermanet et al. \cite{sermanet2018time} introduce a self-supervised representation learning method using time-contrastive networks (TCN) that is invariant to different viewpoints and embodiments. TCN trains a neural network to learn an embedding of each video frame to extract features invariant to context differences, such as the camera angle. By using a triplet loss function, two frames occurring at the same time but with different modalities (i.e., viewpoints) are brought closer together in the embedding space while the frames from distant time-steps but with a visually similar frame are pushed apart (Fig. \ref{embedding}). In order to construct the reward function, Euclidean distance is calculated between the embedding of a demonstration and the embedding of an agent’s camera images. RL techniques are used to optimize the reward function for learning imitation policies. A limitation of this technique is that it requires multi-viewpoint video for training, which is not readily available (e.g., over the Internet).

BC from observation (BCO) \cite{10.5555/3304652.3304697} aims to minimize the amount of post-demonstration environment interactions required for training RL algorithms of prior methods by taking a behavior cloning approach. BCO first learns an inverse dynamics model by letting an agent, who initially follows a random policy, interact with its environment and collect data \cite{edwards2019imitating}. After that, the model is used to infer the missing actions of the expert demonstrations. A BC algorithm is then used to map the states to the inferred actions and solve the problem as a regular IL problem. Using this approach, it is necessary to gather large amounts of data to learn the dynamics model online, especially in high-dimensional problems. 

Subsequent work by \cite{edwards2019imitating} attempts to reduce the number of environment interactions required in BCO by learning a latent forward dynamics model in an offline manner \cite{torabi2019recent}. The underlying assumption in this work is that predictable, though unknown, causes describe the classes of observed state transitions. The goal is to enable an agent to predict and imitate these latent causes \cite{edwards2019imitating}. To achieve this, a latent policy is learned, which estimates the probability that a latent action would be taken in an observed state. They then use a limited number of interactions with the environment to learn a mapping between the real-world actions the agent can take and latent actions identified by the model.

Generative adversarial imitation from observation (GAIfO) \cite{ICML19a-torabi} adapts the GAIL objective to IfO by matching the expert and agent's state-transition distributions. By adopting an adversarial approach, this method can overcome the covariate shift problem encountered in the previous approaches \cite{10.5555/3304652.3304697,edwards2019imitating}. In addition, it is capable of handling demonstrations that are not time-aligned, unlike previous approaches. Using this approach is most successful when the expert and the agent operate in the same environment, under the same dynamics. However, it becomes more challenging to match state-transition distributions when dynamics differ since the expert's state transitions might not even be feasible in the agent's environment \cite{gangwani2022imitation}.

Jaegle et al. \cite{jaegle2021imitation} introduce a non-adversarial IRL from observations approach using likelihood-based generative models. In this method, conditional state transition probabilities are matched between expert and learner.  According to the authors' findings, their approach of matching conditional state transition probabilities tends to focus less on irrelevant differences between the expert and the learner settings than adversarial approaches such as GAIfO, which matches joint state-next-state probabilities. In particular, they argue that conditional state probabilities are less prone to erroneously penalize features that are not present in the demonstrations but lead to correct transitions.

\begin{figure*}[!t]
\centering
\includegraphics[width=6.5in]{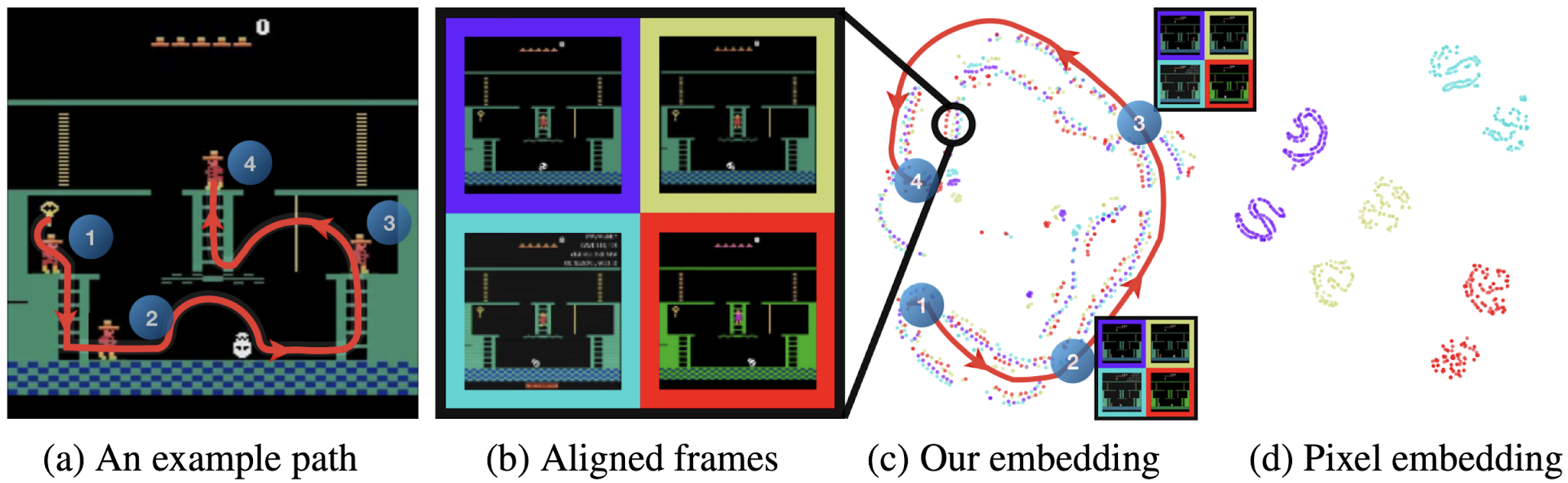}
\caption{For the path in (a), t-SNE projections \cite{van2008visualizing} of trajectories using the proposed embedding (c) and raw pixels (d) are shown \cite{aytar2018playing}. In b, an example frame of MONTEZUMA'S REVENGE is compared using four different domains: the Arcade Learning Environment, and three YouTube videos. Based on the embedding space, it can be seen that the four trajectories are well-aligned.}
\label{youtube}
\end{figure*}

Raychaudhuri et al. \cite{raychaudhuri2021cross} propose a framework that learns state maps across the source and target domain from unpaired, unaligned demonstrations. This approach addresses embodiment, viewpoint, and dynamics mismatch. In order to preserve MDP dynamics during domain transformation, local and global alignments are performed. In local alignment, adversarial training is used to minimize divergence between the state-transition distributions of the true and transferred trajectories. Meanwhile, a learned temporal position function is used to enforce global alignment to ensure that states are placed in a consistent temporal position across the two domains. Finally, given the set of transferred demonstrations, BCO is used to learn the final policy. As with \cite{liu2018imitation}, this method relies on proxy tasks, i.e., expert demonstrations from both domains, which limits its application.

\newcommand{\specialcell}[2][c]{%
  \begin{tabular}[#1]{@{}c@{}}#2\end{tabular}}

\begin{table*}[!htbp]
\centering
\renewcommand\thetable{I}
\caption{{Summary of existing research on imitation learning}}
\label{tbl:AllPapers}
	\resizebox{\textwidth}{0.47\textheight}{\begin{tabular}{ccccccc}
			\hline

\toprule
Ref                                                     & Datasets                                   & Inputs                                        & Learning Type                                                          & \specialcell{Online/\\Offline}              & \specialcell{Online\\Expert}  & Application                                                                                           \\ \midrule
\multicolumn{1}{c}{\cite{sun2017deeply}}              & \multicolumn{1}{c}{Sim}             & \multicolumn{1}{c}{State, Image} & \multicolumn{1}{c}{Interactive IL}                                    & \multicolumn{1}{c}{Online}  & \multicolumn{1}{c}{Yes} & \multicolumn{1}{c}{\specialcell{Robotic Locomotion,\\Dependency Parsing}}                                     \\ \midrule
\multicolumn{1}{c}{\cite{reddy2020sqil}}              & \multicolumn{1}{c}{Sim}             & \multicolumn{1}{c}{State, Image} & \multicolumn{1}{c}{Regularized BC}                    & \multicolumn{1}{c}{Online}  & \multicolumn{1}{c}{No}  & \multicolumn{1}{c}{\specialcell{Car Racing, Atari Games,\\Locomotion Control Tasks}}                                \\ \midrule
\multicolumn{1}{c}{\cite{ross2011reduction}}          & \multicolumn{1}{c}{Sim}             & \multicolumn{1}{c}{State, Image} & \multicolumn{1}{c}{BC\textbackslash DAgger}                         & \multicolumn{1}{c}{Online}  & \multicolumn{1}{c}{Yes} & \multicolumn{1}{c}{Games, Handwriting Recognition}                                                   \\ \midrule
\multicolumn{1}{c}{\cite{kelly2019hg}}                & \multicolumn{1}{c}{Sim,  Real} & \multicolumn{1}{c}{State}              & \multicolumn{1}{c}{BC\textbackslash HG-DAgger}                      & \multicolumn{1}{c}{Online}  & \multicolumn{1}{c}{Yes} & \multicolumn{1}{c}{Autonomous Driving}                                                               \\ \midrule
\multicolumn{1}{c}{\cite{mandlekar2020human}}         & \multicolumn{1}{c}{Sim}             & \multicolumn{1}{c}{State}              & \multicolumn{1}{c}{\specialcell{BC\textbackslash \\Human-gated Interactive IL}}     & \multicolumn{1}{c}{Online}  & \multicolumn{1}{c}{Yes} & \multicolumn{1}{c}{Robotic Manipulation}                                                       \\ \midrule
\multicolumn{1}{c}{\cite{li2022efficient}}            & \multicolumn{1}{c}{Sim}             & \multicolumn{1}{c}{State}              & \multicolumn{1}{c}{Human-in-the-loop RL}                              & \multicolumn{1}{c}{Online}  & \multicolumn{1}{c}{Yes} & \multicolumn{1}{c}{Autonomous Driving}                                                               \\ \midrule
\multicolumn{1}{c}{\cite{Zhang2017query}}             & \multicolumn{1}{c}{Sim}             & \multicolumn{1}{c}{Image}              & \multicolumn{1}{c}{BC\textbackslash SafeDAgger}                     & \multicolumn{1}{c}{Online}  & \multicolumn{1}{c}{Yes} & \multicolumn{1}{c}{Autonomous Driving}                                                               \\ \midrule
\multicolumn{1}{c}{\cite{hoque2021lazydagger}}        & \multicolumn{1}{c}{Sim,  Real} & \multicolumn{1}{c}{State, Image} & \multicolumn{1}{c}{BC\textbackslash LazyDagger}                     & \multicolumn{1}{c}{Online}  & \multicolumn{1}{c}{Yes} & \multicolumn{1}{c}{Robotic Locomotion, Fabric Manipulation}                                    \\ \midrule
\multicolumn{1}{c}{\cite{menda2019ensembledagger}}    & \multicolumn{1}{c}{Sim}             & \multicolumn{1}{c}{State}              & \multicolumn{1}{c}{BC\textbackslash EnsembleDagger}                 & \multicolumn{1}{c}{Online}  & \multicolumn{1}{c}{Yes} & \multicolumn{1}{c}{Inverted Pendulum, Locomotion}                                      \\ \midrule
\multicolumn{1}{c}{\cite{hoque2021thriftydagger}}     & \multicolumn{1}{c}{Sim,  Real} & \multicolumn{1}{c}{State, Image} & \multicolumn{1}{c}{BC\textbackslash ThriftyDAgger}                  & \multicolumn{1}{c}{Online}  & \multicolumn{1}{c}{Yes} & \multicolumn{1}{c}{Peg Insertion, Cable Routing}                                                     \\ \midrule
\multicolumn{1}{c}{\cite{wang2019random}}             & \multicolumn{1}{c}{Sim}             & \multicolumn{1}{c}{State}              & \multicolumn{1}{c}{\specialcell{IL via\\Expert Support Estimation}}           & \multicolumn{1}{c}{Online}  & \multicolumn{1}{c}{No}  & \multicolumn{1}{c}{\specialcell{Robotic Locomotion,\\Autonomous Driving}}                                    \\ \midrule
\multicolumn{1}{c}{\cite{brantley2019disagreement}}   & \multicolumn{1}{c}{Sim}             & \multicolumn{1}{c}{State, Image} & \multicolumn{1}{c}{\specialcell{IL via\\Expert Support Estimation}}           & \multicolumn{1}{c}{Online}  & \multicolumn{1}{c}{No}  & \multicolumn{1}{c}{\specialcell{Atari Games,\\Continuous Control Tasks}}                                            \\ \midrule
\multicolumn{1}{c}{\cite{Bansal2O19ChauffeurNet}}     & \multicolumn{1}{c}{Sim,  Real} & \multicolumn{1}{c}{State}              & \multicolumn{1}{c}{\specialcell{IL via\\Expert Support Estimation}}           & \multicolumn{1}{c}{Online}  & \multicolumn{1}{c}{No}  & \multicolumn{1}{c}{Autonomous Driving}                                                               \\ \midrule
\multicolumn{1}{c}{\cite{wong2022error}}              & \multicolumn{1}{c}{Sim}             & \multicolumn{1}{c}{State}              & \multicolumn{1}{c}{BC}                                & \multicolumn{1}{c}{Online}  & \multicolumn{1}{c}{No}  & \multicolumn{1}{c}{Robotic Manipulation}                                                       \\ \midrule
\multicolumn{1}{c}{\cite{chang2021mitigating}}        & \multicolumn{1}{c}{Sim}             & \multicolumn{1}{c}{State}              & \multicolumn{1}{c}{Offline Imitation Learning}                        & \multicolumn{1}{c}{Offline} & \multicolumn{1}{c}{No}  & \multicolumn{1}{c}{Robotic Locomotion}                                                         \\ \midrule
\multicolumn{1}{c}{\cite{de2019causal}}               & \multicolumn{1}{c}{Sim}             & \multicolumn{1}{c}{State}              & \multicolumn{1}{c}{\specialcell{Causal Graph-Parameterized\\Policy Learning}}        & \multicolumn{1}{c}{Online}  & \multicolumn{1}{c}{Yes} & \multicolumn{1}{c}{\specialcell{Atari Games,\\Robotic Locomotion}}                                            \\ \midrule
\multicolumn{1}{c}{\cite{wen2020fighting}}            & \multicolumn{1}{c}{Sim}             & \multicolumn{1}{c}{State}              & \multicolumn{1}{c}{BC}                                & \multicolumn{1}{c}{Offline} & \multicolumn{1}{c}{No}  & \multicolumn{1}{c}{Robotic Locomotion}                                                         \\ \midrule
\multicolumn{1}{c}{\cite{chuang2022resolving}}        & \multicolumn{1}{c}{Sim}             & \multicolumn{1}{c}{State, Image} & \multicolumn{1}{c}{BC}                                & \multicolumn{1}{c}{Offline} & \multicolumn{1}{c}{No}  & \multicolumn{1}{c}{Robotic Locomotion, Autonomous Driving}                                    \\ \midrule
\multicolumn{1}{c}{\cite{florence2022implicit}}       & \multicolumn{1}{c}{Sim,  Real} & \multicolumn{1}{c}{State, Image} & \multicolumn{1}{c}{Implicit BC}                       & \multicolumn{1}{c}{Offline} & \multicolumn{1}{c}{No}  & \multicolumn{1}{c}{Robotic Manipulation}                                                       \\ \midrule
\multicolumn{1}{c}{\cite{pflueger2019rover}}          & \multicolumn{1}{c}{Sim,  Real} & \multicolumn{1}{c}{State}              & \multicolumn{1}{c}{Maximum Entropy IRL}                               & \multicolumn{1}{c}{Online}  & \multicolumn{1}{c}{No}  & \multicolumn{1}{c}{Path Planning, Gridworld}                                                         \\ \midrule
\multicolumn{1}{c}{\cite{abbeel2004apprenticeship}}   & \multicolumn{1}{c}{Sim}             & \multicolumn{1}{c}{State}              & \multicolumn{1}{c}{Feature Based IRL}                                 & \multicolumn{1}{c}{Online}  & \multicolumn{1}{c}{No}  & \multicolumn{1}{c}{Autonomous Driving, Gridworld}                                                    \\ \midrule
\multicolumn{1}{c}{\cite{ziebart2008maximum}}         & \multicolumn{1}{c}{Real}             & \multicolumn{1}{c}{State}              & \multicolumn{1}{c}{Maximum Entropy IRL}                               & \multicolumn{1}{c}{Online}  & \multicolumn{1}{c}{No}  & \multicolumn{1}{c}{\specialcell{Predicting Driving Behavior,\\Route Recommendation}}                                \\ \midrule
\multicolumn{1}{c}{\cite{ratliff2006maximum}}         & \multicolumn{1}{c}{Real}             & \multicolumn{1}{c}{State}              & \multicolumn{1}{c}{Maximum-margin IRL}                                & \multicolumn{1}{c}{Online}  & \multicolumn{1}{c}{No}  & \multicolumn{1}{c}{Route Planning}                                                                   \\ \midrule
\multicolumn{1}{c}{\cite{bagnell2006boosting}}        & \multicolumn{1}{c}{Sim,  Real} & \multicolumn{1}{c}{State, Image} & \multicolumn{1}{c}{\specialcell{Maximum-margin IRL\textbackslash \\MMPBOOST}}                       & \multicolumn{1}{c}{Online}  & \multicolumn{1}{c}{No}  & \multicolumn{1}{c}{\specialcell{Path Planning, Legged Locomotion,\\Driving Obstacle Detection/Avoidance}}           \\ \midrule
\multicolumn{1}{c}{\cite{ratliff2009learning}}        & \multicolumn{1}{c}{Sim,  Real} & \multicolumn{1}{c}{State, Image} & \multicolumn{1}{c}{Maximum-margin IRL}                         & \multicolumn{1}{c}{Online}  & \multicolumn{1}{c}{No}  & \multicolumn{1}{c}{\specialcell{Footstep Prediction, Grasp Prediction,\\Navigation Task}}                           \\ \midrule
\multicolumn{1}{c}{\cite{syed2007game}}               & \multicolumn{1}{c}{Sim}             & \multicolumn{1}{c}{State}              & \multicolumn{1}{c}{Maximum-margin IRL}                                & \multicolumn{1}{c}{Online}  & \multicolumn{1}{c}{No}  & \multicolumn{1}{c}{Car Driving Game}                                                                 \\ \midrule
\multicolumn{1}{c}{\cite{aghasadeghi2011maximum}}     & \multicolumn{1}{c}{Sim}             & \multicolumn{1}{c}{State}              & \multicolumn{1}{c}{Maximum Entropy IRL}                               & \multicolumn{1}{c}{Online}  & \multicolumn{1}{c}{No}  & \multicolumn{1}{c}{2-D Point Mass Control System}                                                    \\ \midrule
\multicolumn{1}{c}{\cite{kalakrishnan2013learning}}   & \multicolumn{1}{c}{Real}             & \multicolumn{1}{c}{State}              & \multicolumn{1}{c}{Maximum Entropy IRL}                               & \multicolumn{1}{c}{Online}  & \multicolumn{1}{c}{No}  & \multicolumn{1}{c}{Robotic Manipulation}                                                       \\ \midrule
\multicolumn{1}{c}{\cite{boularias2011relative}}      & \multicolumn{1}{c}{Sim}             & \multicolumn{1}{c}{State}              & \multicolumn{1}{c}{Relative Entropy IRL}                              & \multicolumn{1}{c}{Online}  & \multicolumn{1}{c}{No}  & \multicolumn{1}{c}{Car Racing, Gridworld, Game}                                        \\ \midrule
\multicolumn{1}{c}{\cite{wulfmeier2015maximum}}       & \multicolumn{1}{c}{Sim}             & \multicolumn{1}{c}{State}              & \multicolumn{1}{c}{Maximum Entropy Deep IRL}                          & \multicolumn{1}{c}{Online}  & \multicolumn{1}{c}{No}  & \multicolumn{1}{c}{Objectworld, Binaryworld}                                                         \\ \midrule
\multicolumn{1}{c}{\cite{finn2016guided}}             & \multicolumn{1}{c}{Sim,  Real} & \multicolumn{1}{c}{State}              & \multicolumn{1}{c}{Maximum Entropy IRL}                               & \multicolumn{1}{c}{Online}  & \multicolumn{1}{c}{No}  & \multicolumn{1}{c}{Robotic Manipulation}                                                       \\ \midrule
\multicolumn{1}{c}{\cite{ramachandran2007bayesian}}   & \multicolumn{1}{c}{Sim}             & \multicolumn{1}{c}{State}              & \multicolumn{1}{c}{Bayesian IRL}                                      & \multicolumn{1}{c}{Online}  & \multicolumn{1}{c}{No}  & \multicolumn{1}{c}{Random Generated MDPs}                                                            \\ \midrule
\multicolumn{1}{c}{\cite{choi2011map}}                & \multicolumn{1}{c}{Sim}             & \multicolumn{1}{c}{State}              & \multicolumn{1}{c}{\specialcell{Bayesian IRL\textbackslash\\ MAP Inference}}                        & \multicolumn{1}{c}{Online}  & \multicolumn{1}{c}{No}  & \multicolumn{1}{c}{Gridworld, Simplified Car Racing}                                                 \\ \midrule

\end{tabular}%
 
}

\end{table*}

\begin{table*}[!htbp]
\centering
\renewcommand\thetable{I}
\caption{{Summary of existing research on imitation learning (Continued)}}
\label{tbl:AllPapers}
	\resizebox{\textwidth}{0.47\textheight}{\begin{tabular}{ccccccc}

\midrule

\multicolumn{1}{c}{\cite{levine2011nonlinear}}        & \multicolumn{1}{c}{Sim}             & \multicolumn{1}{c}{State}              & \multicolumn{1}{c}{Nonlinear Bayesian IRL}                            & \multicolumn{1}{c}{Online}  & \multicolumn{1}{c}{No}  & \multicolumn{1}{c}{Objectworld,  Highway Driving}                                                    \\ \midrule
\multicolumn{1}{c}{\cite{brown2020safe}}              & \multicolumn{1}{c}{Sim}             & \multicolumn{1}{c}{Image}              & \multicolumn{1}{c}{Bayesian IRL}                                      & \multicolumn{1}{c}{Online}  & \multicolumn{1}{c}{No}  & \multicolumn{1}{c}{Atari Games}                                                                      \\ \midrule
\multicolumn{1}{c}{\cite{chan2021scalable}}           & \multicolumn{1}{c}{Sim,  Real} & \multicolumn{1}{c}{State}              & \multicolumn{1}{c}{Bayesian IRL}                                      & \multicolumn{1}{c}{Offline} & \multicolumn{1}{c}{No}  & \multicolumn{1}{c}{\specialcell{Medical Information Dataset,\\Physics-based Control}} \\ \midrule
\multicolumn{1}{c}{\cite{metelli2021provably}}        & \multicolumn{1}{c}{Sim}             & \multicolumn{1}{c}{Image}              & \multicolumn{1}{c}{IRL}                                               & \multicolumn{1}{c}{Online}  & \multicolumn{1}{c}{Yes} & \multicolumn{1}{c}{\specialcell{Gridworld, Random Generated MDPs,\\Chain MDP}}                                      \\ \midrule
\multicolumn{1}{c}{\cite{lindner2022active}}          & \multicolumn{1}{c}{Sim}             & \multicolumn{1}{c}{State}              & \multicolumn{1}{c}{IRL\textbackslash Active Exploration}                            & \multicolumn{1}{c}{Online}  & \multicolumn{1}{c}{No}  & \multicolumn{1}{c}{\specialcell{Random MDP, Gridworld,\\Chain MDP, Double Chain}}                                   \\ \midrule
\multicolumn{1}{c}{\cite{ho2016generative}}           & \multicolumn{1}{c}{Sim}             & \multicolumn{1}{c}{State}              & \multicolumn{1}{c}{Generative Adversarial IL}                         & \multicolumn{1}{c}{Online}  & \multicolumn{1}{c}{No}  & \multicolumn{1}{c}{Physics-based Control, Robotic Locomotion}                            \\ \midrule
\multicolumn{1}{c}{\cite{fu2017learning}}             & \multicolumn{1}{c}{Sim}             & \multicolumn{1}{c}{State}              & \multicolumn{1}{c}{Adversarial IRL}                                   & \multicolumn{1}{c}{Online}  & \multicolumn{1}{c}{No}  & \multicolumn{1}{c}{\specialcell{Random Generated MDPs,\\Continuous Control Tasks}}                                  \\ \midrule
\multicolumn{1}{c}{\cite{kostrikov2018discriminator}} & \multicolumn{1}{c}{Sim}             & \multicolumn{1}{c}{State}              & \multicolumn{1}{c}{Adversarial IRL}                                   & \multicolumn{1}{c}{Online}  & \multicolumn{1}{c}{No}  & \multicolumn{1}{c}{Robotic Locomotion and Manipulation}                                        \\ \midrule
\multicolumn{1}{c}{\cite{jeon2018bayesian}}           & \multicolumn{1}{c}{Sim}             & \multicolumn{1}{c}{State}              & \multicolumn{1}{c}{Bayes-GAIL}                                        & \multicolumn{1}{c}{Online}  & \multicolumn{1}{c}{No}  & \multicolumn{1}{c}{Robotic Locomotion}                                                         \\ \midrule
\multicolumn{1}{c}{\cite{sasaki2018sample}}           & \multicolumn{1}{c}{Sim}             & \multicolumn{1}{c}{State}              & \multicolumn{1}{c}{IRL}                                               & \multicolumn{1}{c}{Online}  & \multicolumn{1}{c}{No}  & \multicolumn{1}{c}{Robotic Locomotion}                                                         \\ \midrule
\multicolumn{1}{c}{\cite{dadashi2020primal}}          & \multicolumn{1}{c}{Sim}             & \multicolumn{1}{c}{State, Image} & \multicolumn{1}{c}{Adversarial IRL}                                   & \multicolumn{1}{c}{Online}  & \multicolumn{1}{c}{No}  & \multicolumn{1}{c}{Robotic Locomotion and Hand Manipulation}                              \\ \midrule
\multicolumn{1}{c}{\cite{li2017infogail}}             & \multicolumn{1}{c}{Sim}             & \multicolumn{1}{c}{State, Image} & \multicolumn{1}{c}{\specialcell{Generative Adversarial IL\textbackslash\\InfoGAIL}}                & \multicolumn{1}{c}{Online}  & \multicolumn{1}{c}{No}  & \multicolumn{1}{c}{\specialcell{Synthetic 2D Example,\\Autonomous Highway Driving}}                                 \\ \midrule
\multicolumn{1}{c}{\cite{liu2018imitation}}           & \multicolumn{1}{c}{Sim,  Real} & \multicolumn{1}{c}{Image}              & \multicolumn{1}{c}{Imitation from Observation}                        & \multicolumn{1}{c}{Online}  & \multicolumn{1}{c}{No}  & \multicolumn{1}{c}{Robotic Manipulation}                                                       \\ \midrule
\multicolumn{1}{c}{\cite{sermanet2018time}}           & \multicolumn{1}{c}{Sim,  Real} & \multicolumn{1}{c}{Image}              & \multicolumn{1}{c}{IfO\textbackslash Self-supervised Learning}        & \multicolumn{1}{c}{Online}  & \multicolumn{1}{c}{No}  & \multicolumn{1}{c}{Robotic Manipulation, Human Pose Imitation}                                 \\ \midrule
\multicolumn{1}{c}{\cite{10.5555/3304652.3304697}}    & \multicolumn{1}{c}{Sim}             & \multicolumn{1}{c}{State}              & \multicolumn{1}{c}{BC from Observation}               & \multicolumn{1}{c}{Online}  & \multicolumn{1}{c}{No}  & \multicolumn{1}{c}{Physics-based Control, Robotic Locomotion}                            \\ \midrule
\multicolumn{1}{c}{\cite{edwards2019imitating}}       & \multicolumn{1}{c}{Sim}             & \multicolumn{1}{c}{State, Image} & \multicolumn{1}{c}{Imitation from Observation}                        & \multicolumn{1}{c}{Online}  & \multicolumn{1}{c}{No}  & \multicolumn{1}{c}{Physics-based Control, CoinRun Game}                                        \\ \midrule
\multicolumn{1}{c}{\cite{ICML19a-torabi}}             & \multicolumn{1}{c}{Sim}             & \multicolumn{1}{c}{State, Image} & \multicolumn{1}{c}{\specialcell{Generative Adversarial Imitation\\from Observation}} & \multicolumn{1}{c}{Online}  & \multicolumn{1}{c}{No}  & \multicolumn{1}{c}{Physics-based Control, Robotic Locomotion}                            \\ \midrule
\multicolumn{1}{c}{\cite{gangwani2022imitation}}      & \multicolumn{1}{c}{Sim}             & \multicolumn{1}{c}{State}              & \multicolumn{1}{c}{Imitation from Observation}                        & \multicolumn{1}{c}{Online}  & \multicolumn{1}{c}{No}  & \multicolumn{1}{c}{Robotic Locomotion}                                                         \\ \midrule
\multicolumn{1}{c}{\cite{jaegle2021imitation}}        & \multicolumn{1}{c}{Sim}             & \multicolumn{1}{c}{State}              & \multicolumn{1}{c}{IRL from Observations}                             & \multicolumn{1}{c}{Online}  & \multicolumn{1}{c}{No}  & \multicolumn{1}{c}{Robotic Locomotion}                                              
                                                               \\ \midrule
\multicolumn{1}{c}{\cite{raychaudhuri2021cross}}      & \multicolumn{1}{c}{Sim}             & \multicolumn{1}{c}{State}              & \multicolumn{1}{c}{Cross-domain IL from Observations}                & \multicolumn{1}{c}{Offline} & \multicolumn{1}{c}{No}  & \multicolumn{1}{c}{Physics-based Control, Robotic Locomotion}                            \\ \midrule

\multicolumn{1}{c}{\cite{aytar2018playing}}           & \multicolumn{1}{c}{Sim}             & \multicolumn{1}{c}{Image}              & \multicolumn{1}{c}{Imitation from Observation}                        & \multicolumn{1}{c}{Online}  & \multicolumn{1}{c}{No}  & \multicolumn{1}{c}{Atari Games}                                                                      \\ \midrule
\multicolumn{1}{c}{\cite{brown2019extrapolating}}     & \multicolumn{1}{c}{Sim}             & \multicolumn{1}{c}{State, Image} & \multicolumn{1}{c}{IRL from Observations}            & \multicolumn{1}{c}{Online}  & \multicolumn{1}{c}{No}  & \multicolumn{1}{c}{Atari Games, Robotic Locomotion}                                            \\ \midrule
\multicolumn{1}{c}{\cite{zhang2022selfd}}             & \multicolumn{1}{c}{Sim,  Real} & \multicolumn{1}{c}{Image}              & \multicolumn{1}{c}{Conditional IL from Observations}                  & \multicolumn{1}{c}{Offline} & \multicolumn{1}{c}{No}  & \multicolumn{1}{c}{Autonomous Driving}                                                               \\ \midrule
\multicolumn{1}{c}{\cite{wu2019imitation}}            & \multicolumn{1}{c}{Sim}             & \multicolumn{1}{c}{State}              & \multicolumn{1}{c}{\specialcell{Importance Weighting IL,\\GAIL}}                     & \multicolumn{1}{c}{Online}  & \multicolumn{1}{c}{No}  & \multicolumn{1}{c}{Robotic Locomotion}                                                         \\ \midrule
\multicolumn{1}{c}{\cite{sasaki2020behavioral}}       & \multicolumn{1}{c}{Sim}             & \multicolumn{1}{c}{State}              & \multicolumn{1}{c}{BC from Noisy Demonstrations}      & \multicolumn{1}{c}{Offline} & \multicolumn{1}{c}{No}  & \multicolumn{1}{c}{Robotic Locomotion}                                                         \\ \midrule
\multicolumn{1}{c}{\cite{wang2021learning}}           & \multicolumn{1}{c}{Sim}             & \multicolumn{1}{c}{State, Image} & \multicolumn{1}{c}{Weighted GAIL}                                     & \multicolumn{1}{c}{Online}  & \multicolumn{1}{c}{No}  & \multicolumn{1}{c}{Atari Games, Robotic Locomotion}                                            \\ \midrule
\multicolumn{1}{c}{\cite{kim2021demodice}}            & \multicolumn{1}{c}{Sim}             & \multicolumn{1}{c}{State}              & \multicolumn{1}{c}{Weighted BC}                                       & \multicolumn{1}{c}{Offline} & \multicolumn{1}{c}{No}  & \multicolumn{1}{c}{Robotic Locomotion}                                                         \\ \midrule
\multicolumn{1}{c}{\cite{beliaev2022imitation}}       & \multicolumn{1}{c}{Sim}             & \multicolumn{1}{c}{State}              & \multicolumn{1}{c}{BC}                                & \multicolumn{1}{c}{Offline} & \multicolumn{1}{c}{No}  & \multicolumn{1}{c}{\specialcell{MiniGrid Environments, Robotic Manipulation,\\Chess Game-endings}}                  \\ \midrule
\multicolumn{1}{c}{\cite{chae2022robust}}             & \multicolumn{1}{c}{Sim}             & \multicolumn{1}{c}{State}              & \multicolumn{1}{c}{GAIL with Modified Objective}                      & \multicolumn{1}{c}{Online}  & \multicolumn{1}{c}{No}  & \multicolumn{1}{c}{Robotic Locomotion}                                                         \\ \midrule
\multicolumn{1}{c}{\cite{stadie2017third}}            & \multicolumn{1}{c}{Sim}             & \multicolumn{1}{c}{State}              & \multicolumn{1}{c}{GAIL}                                              & \multicolumn{1}{c}{Online}  & \multicolumn{1}{c}{No}  & \multicolumn{1}{c}{Physics-based Control}                                                      \\ \midrule
\multicolumn{1}{c}{\cite{zakka2022xirl}}              & \multicolumn{1}{c}{Sim,  Real} & \multicolumn{1}{c}{Image}              & \multicolumn{1}{c}{Cross-embodiment IRL}                              & \multicolumn{1}{c}{Online}  & \multicolumn{1}{c}{No}  & \multicolumn{1}{c}{Robotic Manipulation}                                                       \\ \midrule
\multicolumn{1}{c}{\cite{fickinger2021cross}}         & \multicolumn{1}{c}{Sim}             & \multicolumn{1}{c}{State}              & \multicolumn{1}{c}{IL via Optimal Transport}                          & \multicolumn{1}{c}{Online}  & \multicolumn{1}{c}{No}  & \multicolumn{1}{c}{\specialcell{Physics-based Control, Robotic Locomotion,\\2D Maze Navigation}}        \\ \midrule
\multicolumn{1}{c}{\cite{kim2020domain}}              & \multicolumn{1}{c}{Sim}             & \multicolumn{1}{c}{State}              & \multicolumn{1}{c}{BC}                                & \multicolumn{1}{c}{Offline} & \multicolumn{1}{c}{No}  & \multicolumn{1}{c}{Robotic Manipulation, Physics-based Control}    \\ \bottomrule

\end{tabular}%
 
}

\end{table*}

Aytar et al. \cite{aytar2018playing} present a novel self-supervised framework for learning to play hard exploration Atari games without ever being explicitly exposed to the Atari environment by watching YouTube videos (Fig. \ref{youtube}). Learning from YouTube videos poses several challenges due to the existence of domain-specific variations (e.g., in color or resolution), and a lack of frame-by-frame alignment. To address these challenges, they first use self-supervised classification tasks, constructed over both vision and audio to map unaligned videos from multiple sources to a common representation. Then, a single YouTube video is embedded in this representation, and a sequence of checkpoints are placed along the embedding. This is used to create a reward function that encourages the agent to imitate human gameplay. During policy training, the agent is rewarded only when it reaches these checkpoints.

Brown et al. \cite{brown2019extrapolating} introduce an IRL from observation technique for extrapolating the expert's intent from suboptimal ranked demonstrations. This work aims to improve the performance over a suboptimal expert in high-dimensional tasks by inferring the expert’s intentions. They learn a state-based reward function such that a greater total return is assigned to higher-ranked trajectories. Utilizing ranking to construct a reward function in this way enables identifying features that are correlated with rankings, allowing for potentially better-than-demonstrator performance. Given the learned reward function, RL is used to optimize a policy.

Utilizing large amounts of navigation data from YouTube, \cite{zhang2022selfd} proposes a framework for learning scalable driving. First, a model is trained on a small labeled dataset to map monocular images to Bird’s Eye View (BEV), facilitating learning from the unconstrained nature of YouTube videos (e.g. in viewpoints or camera parameters). Since many publicly available driving datasets include action labels, this assumption is reasonable. This trained model is used to generate pseudo-labels over a large unlabeled dataset. Lastly, a generalized policy is trained on the pseudo-labeled dataset and fine-tuned on the clean labels of the small labeled dataset.

\section{Challenges and Limitations}
\subsection{Imperfect Demonstrations}
A common assumption in IL methods is that the demonstrations will be optimal, performed by an expert demonstrator \cite{ravichandar2020recent}. However, this assumption is too restrictive when it comes to learning from demonstrations in a variety of cases \cite{ravichandar2020recent}. Firstly, it can be difficult to obtain large numbers of high-quality demonstrations from human experts \cite{yang2021trail, xiang2019task}. In many real-world tasks, this would be impossible for humans due to the amount of time and effort required. Additionally, humans are prone to making mistakes for various reasons, such as the presence of distractions, or limited observability of the environment \cite{wu2019imitation, sasaki2020behavioral}. Secondly, it is necessary to leverage the scale and diversity of crowd-sourced datasets to learn robust effective IL policies \cite{ramrakhya2022habitat}. However, a crowd-sourced dataset will inevitably have a wide range of behavior optimality since it is collected from users with varying levels of expertise.

The naive solution to imperfect demonstrations would be to discard the non-optimal ones. However, this screening process is often impractical since it requires significant human effort \cite{sasaki2020behavioral}. Therefore, researchers have been increasingly interested in developing methods that can learn from imperfect demonstrations.

Wu et al. \cite{wu2019imitation} present two general approaches to address imperfect demonstrations by utilizing both confidence-scored and unlabeled data: two-step importance weighting IL (2IWIL) and generative adversarial IL with imperfect demonstration and confidence (IC-GAIL). Both approaches assume that a fraction of demonstrations are annotated with confidence scores (i.e. the probability that a given trajectory is optimal). 2IWIL is a two-step approach that first uses a semi-supervised classifier to generate confidence scores for the unlabeled demonstrations, and then performs standard GAIL with reweighted distribution \cite{kim2021demodice}. To avoid error accumulation in two steps, IC-GAIL forgoes learning a classifier and performs occupancy measure matching with unlabeled demonstrations.

Sasaki et al. \cite{sasaki2020behavioral} propose an offline BC algorithm to learn from noisy demonstrations, obtained from a noisy expert, without any screening or annotations associated with the non-optimal demonstrations.  The key idea is to leverage the learned policy to reweight the samples in the next iteration of weighted BC. The noisy expert action distribution is assumed to be a weighted mixture of two distributions: the action distribution of an optimal expert and a non-optimal one. The goal is to change the weights so that the noisy expert action distribution mode gets closer to the optimal expert action distribution mode. This is achieved by reusing the old policy (i.e. the policy optimized in the previous iteration) as the weights for action samples in the weighted BC objective. However, this approach only converges to the optimal policy when optimal demonstrations constitute the majority of the data.

Wang et al. \cite{wang2021learning} investigate how to weight imperfect demonstrations in GAIL without requiring auxiliary information from an oracle. An automatic weight prediction method is proposed to assess the quality and significance of each demonstration for training. They demonstrate that the weight can be accurately estimated using both the discriminator and the agent policy in GAIL. In the training procedure, the weight estimation is conducted first to determine weight for each demonstration. Using weighted GAIL, the agent policy is then trained with weighted demonstrations. These two procedures interact alternately and are optimized as a whole.

Kim et al. \cite{kim2021demodice} aim to overcome the distributional shift problem caused by the lack of sufficient expert demonstrations by using supplementary imperfect demonstrations with unknown optimality levels. They regularize a distribution-matching objective of IL by a KL divergence between the agent distribution and a mixture of expert and imperfect distributions. An optimal state-action distribution of this regularized objective is obtained using a dual-program technique \cite{lee2021optidice}. Given the optimal state-action distribution, the expert policy is extracted by performing weighted BC.

\begin{figure*}[!t]
\centering
\includegraphics[width=7in]{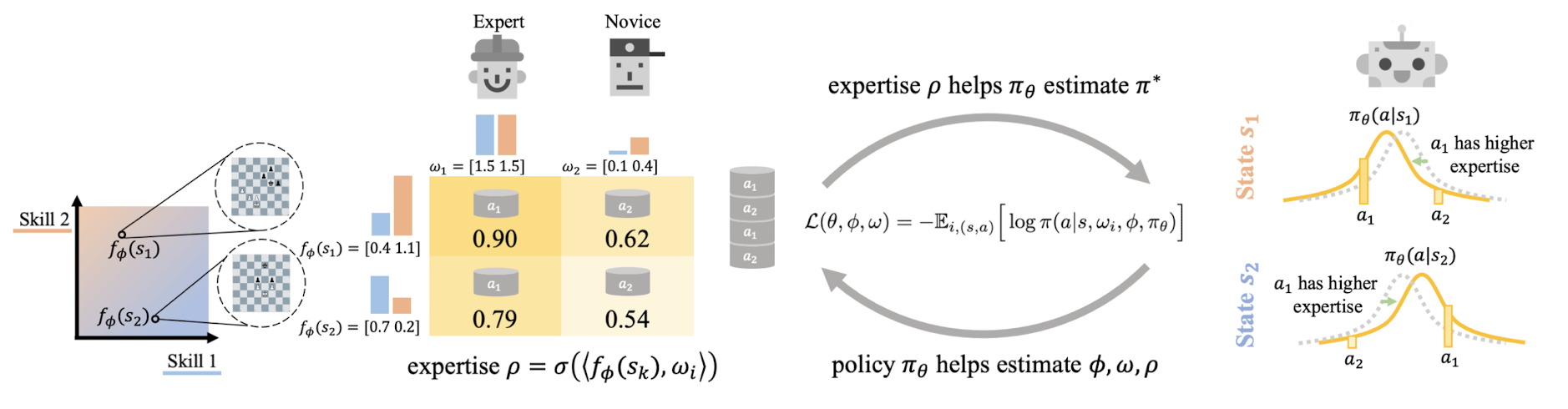}
\caption{IL by estimating expertise of demonstrators \cite{beliaev2022imitation}. Left: Skills associated with states are encoded by state embeddings. Middle: Demonstrators' expertise, $\rho$, at a particular state is determined by the state embedding and the demonstrator embedding. Right: By utilizing the expertise levels, the model improves the learned policy, thereby improving the estimation of the state/demonstrator embeddings and the expertise level.}
\hfil
\label{expertise}
\end{figure*}

Beliaev et al. \cite{beliaev2022imitation} introduce IL by Estimating Expertise of Demonstrators (ILEED). It leverages information about demonstrators’ identities to infer their expertise levels in an unsupervised manner. Each demonstrator is assigned a state-dependent expertise value, which indicates which demonstrators perform better in specific states, allowing them to combine their strengths in different states. ILEED develops and optimizes a joint model over a learned policy and expertise levels. As a result, the model is able to learn from the optimal behavior of each demonstrator and filter out the suboptimal behavior. Expertise levels are modeled by the inner product of two embeddings: an state embedding, and a demonstrator embedding. Each dimension of the embeddings vector corresponds to a latent skill, with the state embedding having a weighting of how relevant that skill is in acting correctly at that state and the demonstrator embedding representing how adept the demonstrator is at that skill (Fig. \ref{expertise}). 

\subsection{Domain Discrepancies}
The majority of prior research assumes that the expert and the agent operate under the same state and action space \cite{fickinger2021cross}. This assumption makes it easier to manually specify one-to-one correspondences between the actions of the expert and the agent. However, this will restrict the application of these algorithms to simple scenarios in which expert demonstrations come from the agent domain. In recent years, there has been increased interest in IL under a more relaxed and realistic assumption: given demonstrations of a task in the expert domain, train the agent to learn to perform the task optimally in its own domain \cite{fickinger2021cross}. This relaxed setting facilitates the collection of demonstrations by removing the requirement for in-domain expert demonstrations, improving the efficiency and scalability of IL. A variety of solutions have been proposed to address three main types of domain discrepancies: dynamics \cite{gangwani2022imitation,chae2022robust}, viewpoint \cite{stadie2017third,sermanet2018time}, and embodiment mismatch \cite{zakka2022xirl,fickinger2021cross, 6942235}. 

The transfer of knowledge between different domains in IL research often involves learning a mapping between the state-action spaces. Recent works \cite{liu2018imitation,sermanet2018time} utilize paired and time-aligned demonstrations from both domains (expert and agent) to learn a state mapping or encoding to a domain invariant feature space. Following this, they perform an RL step to learn the final policy on the given task. These studies are limited in their application due to the limited availability of paired demonstrations and the high cost of RL procedures \cite{kim2020domain}. To overcome these limitations, Kim et al. \cite{kim2020domain} propose a general framework to learn state and action maps from unpaired and unaligned demonstrations while having access to an online expert. In addition, they eliminate the need for an expensive RL step by leveraging the action map to perform zero-shot imitation. The work of \cite{raychaudhuri2021cross} extends \cite{kim2020domain} to the imitation form observation setting and also eliminates the need for an online expert in \cite{kim2020domain}. All these methods rely on proxy tasks, which limits their applicability in real-world scenarios. Stadie et al. \cite{stadie2017third} propose an adversarial framework for viewpoint-agnostic imitation that uses a discriminator to distinguish data coming from different viewpoints and maximizes domain confusion without proxy tasks. Zakka et al. \cite{zakka2022xirl} adopt a goal-driven approach that focuses on imitating task progress rather than matching fine-grained structural details. Several of these approaches have already been discussed in previous sections. The following is a detailed discussion of a few of the most recent methods.

\begin{figure}[!t]
\centering
\includegraphics[width=3in]{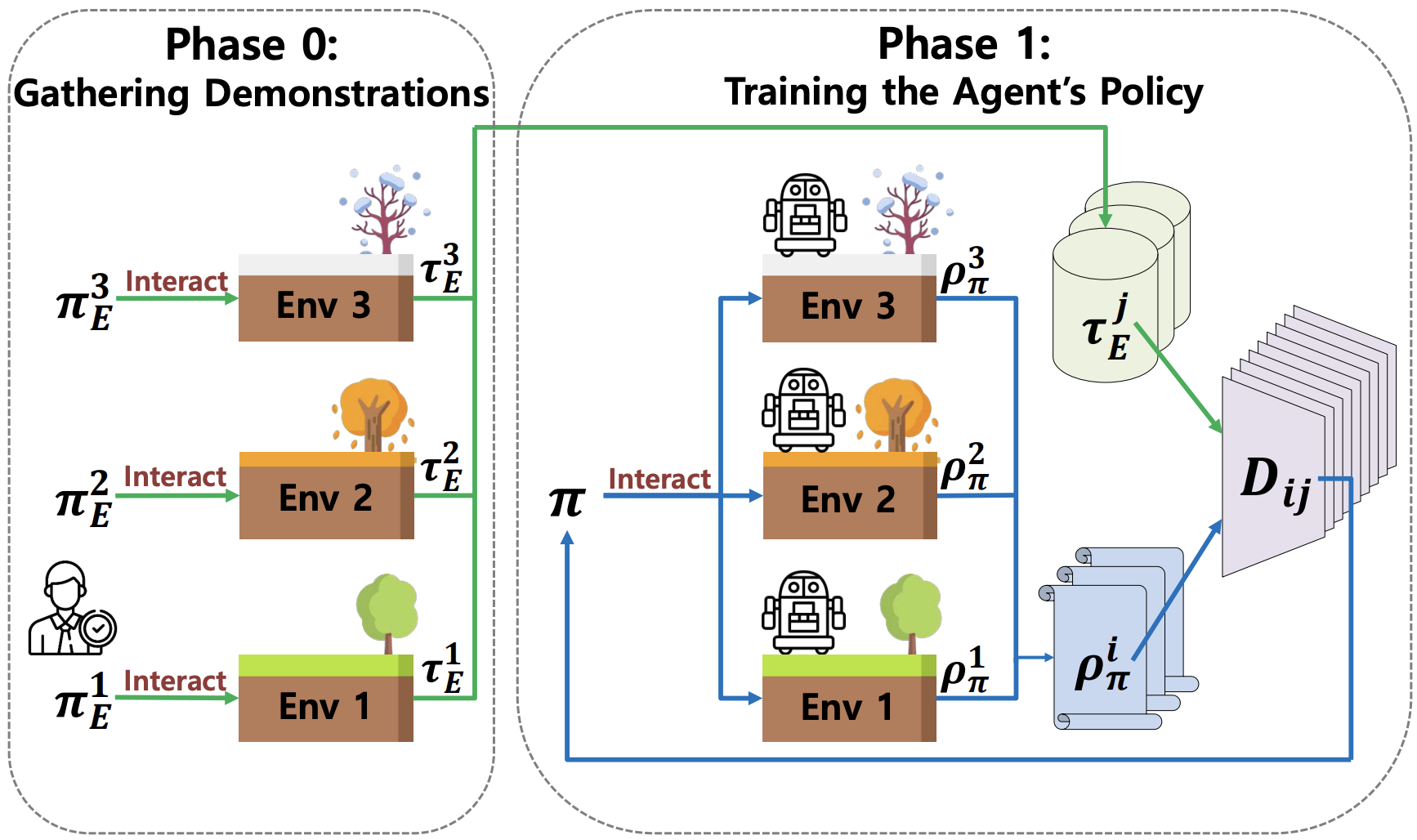}
\caption{IL against variations in environment dynamics \cite{chae2022robust}. Blue represents the flow of policy samples, while green represents the flow of expert demonstrations.}
\label{dynamix}
\end{figure}

Chae et al. \cite{chae2022robust} provide a framework for learning policies that can perform well under perturbed environment dynamics. The objective is to train a policy that is robust to continuous dynamics variation, using only a few samples from the continuum of environment dynamics. The sampled environments are used during both the demonstration collection phase and the policy interaction phase (Fig. \ref{dynamix}). The problem is then formulated as the minimization of the weighted average of Jensen-Shannon divergences between the multiple expert policies and the agent policy.

Cross-embodiment IRL (XIRL) \cite{zakka2022xirl} attempts to extract an agent-invariant definition of a task from videos of different agents executing the same task differently due to embodiment differences. XIRL uses temporal cycle consistency (TCC) to learn visual embeddings, which identify key moments in videos of varying lengths and cluster them to encode task progression. In order to learn an embodiment-invariant reward function, XIRL uses the distance from a single goal state in the TCC embedding space. This method can be applied to any number of embodiments or experts (regardless of their skill level) since it does not require the manual pairing of video frames between the expert and the learner.

Fickinger et al. \cite{fickinger2021cross} examine how expert demonstrations can be used to train an imitator agent with a different embodiment without relying on explicit cross-domain latent space \cite{zakka2022xirl} or resorting to any form of proxy tasks \cite{raychaudhuri2021cross,sermanet2018time,liu2018imitation}. Instead, they use the Gromov-Wasserstein distance between state-action occupancies of the expert and the agent to find isometric transformations that preserve distance measures between the two domains. Given trajectories from the expert and the agent domain, pseudo-rewards are computed based on the degree to which distances from a state to its neighbors in the agent domain are preserved in the expert domain. Using these pseudo-rewards, the policy is optimized using an RL algorithm.

\section{Opportunities and Future Work}

This survey paper provides a comprehensive overview of the field of IL, exploring its algorithms, categorizations, developments, and challenges. The paper starts by presenting a categorization of IL algorithms, identifying two general learning approaches, namely BC and IRL, and discussing their relative benefits and limitations. Additionally, the paper highlights the benefits of integrating adversarial training into IL and evaluates the current progress in the AIL field. The paper also introduces a novel technique called IfO that aims to learn from state-only demonstrations.

Through the examination of various IL algorithms, we have gained valuable insights into their strengths and limitations and identified some of the key challenges and opportunities for future research. One of the significant challenges across all categories of IL approaches is the need to collect diverse and large-scale demonstrations, which is crucial for training a generalizable policy that can be applied in the real world \mbox{\cite{ramrakhya2022habitat}}. However, this poses a challenge, as readily available demonstration resources such as online videos present additional difficulties such as the varying levels of expertise among the demonstrators.

Another challenge in IL research is developing methods that enable agents to learn across domains with differences in dynamics, viewpoint, and embodiment. Overcoming these challenges is essential if we are to teach agents to learn from experts effectively and apply the insights from IL research to real-world scenarios. Therefore, future research should focus on developing algorithms that can learn from imperfect demonstrations, extract useful information, and enable cross-domain learning.
Despite these challenges, the field of IL presents exciting opportunities for future research. As the field of AI continues to evolve and mature, we believe that IL will play a critical role in enabling agents to learn from demonstrations, adapt to new tasks and environments, and ultimately achieve more advanced levels of intelligence, paving the way for real-world applications of AI.

\section*{Acknowledgments}
This research was partially supported by the Australian Research Council’s Discovery Projects funding scheme (project DP190102181 and DP210101465).

\bibliography{Survey.bib}{}

\bibliographystyle{IEEEtran}

\end{document}